\documentclass{article} 

\usepackage[accepted]{icml2024}
\usepackage{icml2024,times}

\usepackage{amsmath,amsfonts,bm}

\def\eqref#1{equation~\ref{#1}}

\def\1{\bm{1}}

\DeclareMathAlphabet{\mathsfit}{\encodingdefault}{\sfdefault}{m}{sl}
\SetMathAlphabet{\mathsfit}{bold}{\encodingdefault}{\sfdefault}{bx}{n}

\usepackage[colorlinks=true, citecolor=teal]{hyperref}
\usepackage{url}
\definecolor{contentsColor}{RGB}{0,20,115}
\hypersetup{linkcolor={contentsColor}}
\usepackage{colortbl}

\usepackage{booktabs}       
\usepackage{amsfonts}       
\usepackage{nicefrac}       
\usepackage{microtype}      
\usepackage[inkscapeformat=png]{svg}

\definecolor{LightGreen}{rgb}{0.93,0.98,0.96}
\definecolor{Green}{rgb}{0.0,0.5,0.0}

\usepackage{graphicx}
\usepackage{amsmath}
\usepackage{amsthm}
\usepackage{amsmath}
\usepackage{xspace}
\usepackage{wrapfig}
\usepackage{lipsum}
\usepackage{graphicx}
\usepackage{amsmath}
\usepackage{amssymb}
\usepackage{booktabs}
\usepackage{multirow}
\usepackage{makecell}
\usepackage{bbm}
\usepackage{enumitem}
\usepackage{multirow}
\usepackage{bm}
\usepackage{bbm}
\usepackage{subcaption}
\usepackage{siunitx}
\usepackage{algorithm}  
\usepackage{graphicx}
\usepackage[T1]{fontenc}
\newcommand{\midsepremove}{\aboverulesep = 0mm \belowrulesep = 0mm}

\usepackage{mathtools}
\newtheorem{theorem}{Theorem}

\newtheorem{definition}{Definition}[section]

\newcommand{\ie}{\textit{i}.\textit{e},\ }
\newcommand{\eg}{\textit{e}.\textit{g}.,\ }

\usepackage{etoc}
\etocdepthtag.toc{mtchapter}
\etocsettagdepth{mtchapter}{subsection}
\etocsettagdepth{mtappendix}{none}
\usepackage[textsize=tiny]{todonotes}

\icmltitlerunning{Mitigating Label Noise on Graphs via Topological Sample Selection}

\begin{document}

\twocolumn[

\icmltitle{Mitigating Label Noise on Graphs via Topological Sample Selection}

\begin{icmlauthorlist}
\icmlauthor{Yuhao Wu}{usyd}
\icmlauthor{Jiangchao Yao}{SJTU,SJlab}
\icmlauthor{Xiaobo Xia}{usyd}
\icmlauthor{Jun Yu}{ustc}
\icmlauthor{Ruxin Wang}{Ali}
\icmlauthor{Bo Han}{hkbu}
\icmlauthor{Tongliang Liu}{usyd}
\end{icmlauthorlist}

\icmlaffiliation{usyd}{Sydney AI Center, The University of Sydney}
\icmlaffiliation{SJTU}{CMIC, Shanghai Jiao Tong University}
\icmlaffiliation{SJlab}{Shanghai AI Laboratory}
\icmlaffiliation{hkbu}{TMLR Group, Department of Computer Science, Hong Kong Baptist University}
\icmlaffiliation{ustc}{University of Science and Technology of China}
\icmlaffiliation{Ali}{Alibaba Group}

\icmlcorrespondingauthor{Jiangchao Yao}{Sunarker@sjtu.edu.cn}
\icmlcorrespondingauthor{Tongliang Liu}{tongliang.liu@sydney.edu.au}

\icmlkeywords{Machine Learning, ICML}

\vskip 0.3in
]

\printAffiliationsAndNotice{}  

\begin{abstract}

Despite the success of the carefully-annotated benchmarks, the effectiveness of existing graph neural networks (GNNs) can be considerably impaired in practice when the real-world graph data is noisily labeled.  Previous explorations in sample selection have been demonstrated as an effective way for robust learning with noisy labels, however, the conventional studies focus on i.i.d data, and when moving to non-iid graph data and GNNs, two notable challenges remain: (1) nodes located near topological class boundaries are very informative for classification but cannot be successfully distinguished by the heuristic sample selection. (2) there is no available measure that considers the graph topological information to promote sample selection in a graph. To address this dilemma, we propose a \textit{Topological Sample Selection} (TSS) method that boosts the informative sample selection process in a graph by utilising topological information. We theoretically prove that our procedure minimizes an upper bound of the expected risk under target clean distribution, and experimentally show the superiority of our method compared with state-of-the-art baselines. Our implementation is available at \url{https://github.com/tmllab/2024_ICML_TSS}.

\end{abstract}

\vspace{-15pt}
\section{Introduction}

Noisy labels ubiquitous in real-world applications~\citep{deng2020sub,mirzasoleiman2020coresets,gao2022missdag,yao2023latent,huang2023machine,chen2024towards,wu2023making} inevitably impair the learning efficiency and the generalization robustness of deep neural networks (DNNs)~\citep{liu2015classification,rolnick2017deep,nguyen2019self,yuan2024early}. 
It becomes exacerbated on the graph data, as the noise influence can be propagated along the topological edges, unlike the independent and identically distributed (i.i.d.) data in the forms of image~\citep{mirzasoleiman2020coresets,chen2019understanding,frenay2013classification,thulasidasan2019combating,nt2019learning,wei2021smooth,chengclass,berthon2021confidence}. Combating the degeneration of GNNs on the noisily labeled graph then emerges as a non-negligible problem, drawing more attention from the research community~\citep{dai2021nrgnn,li2021unified,du2021pi,yuan2023alex,yuan2023learning,xia2023gnn,yao2021instance,yao2020dual,lin2023cs}.

Sample selection has been demonstrated as a promising way to deal with label noise on i.i.d. data~\citep{han2018co,jiang2018mentornet,zhou2020robust,yuan2023late,yao2023better,li2024instant}, due to its simplicity and effectiveness in isolating incorrectly labeled samples. It builds upon the memorization effect that clean samples will be learned before mislabeled samples, which allows designing strategies of extracting clean samples corresponding to the predictions of the trained model \ie small loss trick or high prediction confidence~\citep{arpit2017closer,cheng2020learning,northcutt2021confident}. The extracted samples are more likely clean and thus will lead a classifier to a clean data regime, thereby mitigating the negative impact posed by corrupted labels.

\begin{figure} 
\centering
\includegraphics[width=1\linewidth]{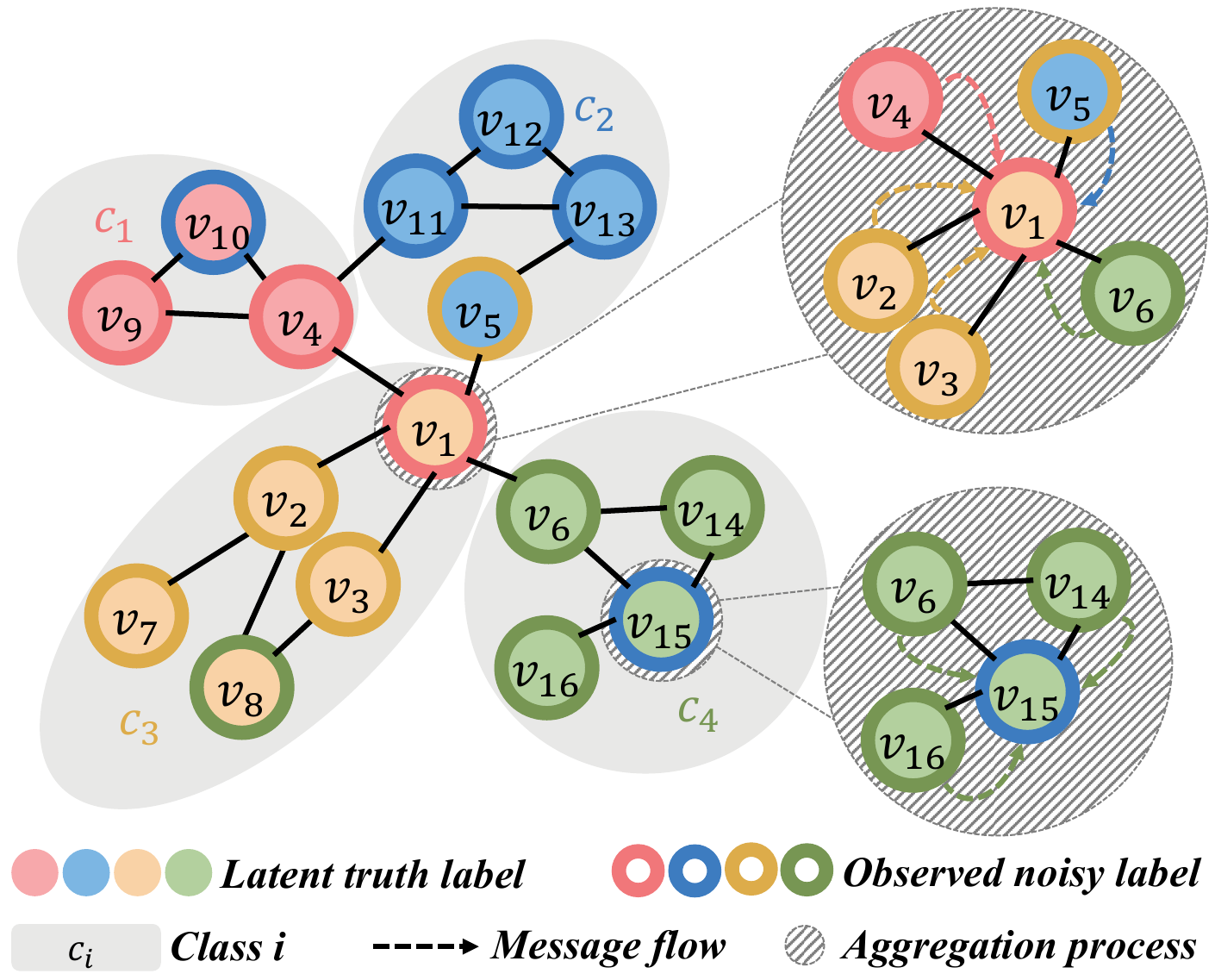}
\vspace{-10pt}
\caption{Illustration of noisily labeled nodes with different topological structures. $v_{1}$ is a mislabeled node located near class boundaries while $v_{15}$ is a mislabeled node far away from class boundaries.}
\vspace{-10pt}
\label{into}
\end{figure}

The straightforward application of such sample selection methods on noisily labeled graph data does not show promise as usual due to the neglect of the important topological information on a graph. As illustrated in Figure~\ref{into},  nodes located near topological class boundaries are much informative compared to nodes located far from topological class boundaries, as they may link nodes from diverse classes~\citep{brandes2001faster,barthelemy2004betweenness,freeman1977set,zhu2020beyond,wu2024unraveling}.
However, those boundary-near clean nodes are harder to learn and identify in noisily labeled graphs compared with the clean nodes that are far from boundaries. Since boundary-near nodes are often of a small proportion and lose discriminative information due to the aggregation from the heterogeneous neighbours in GNNs, they are often entangled with mislabeled nodes in the procedures of sample selection~\citep{bai2021me,wei2023clnode}. Besides, there is a scarcity of a method that considers the topological characteristic within a noisily labeled graph to promote informative sample selection.

To address this dilemma, we propose a robust \textit{Topological Sample Selection} (TSS) method, specifically designed to progressively select informative samples in a noisily labeled graph. Initially, TSS extracts nodes located far from class boundaries and these extracted nodes are more likely to be clean. These initially extracted nodes make the model in TSS learn a clean pattern, subsequently facilitating the extraction of informative clean nodes near class boundaries, where those nodes are often entangled with incorrect labels. Identifying the topological positions of nodes is based on a proposed \textit{Class-conditional Betweenness Centrality} (CBC) measure, which quantifies the heterogeneous message passing between different classes. This measurement is inspired by the \textit{random-walk} technique applied to graph structures~\citep{nikolentzos2020random, kang2012fast} rather than GNN predictions. As a result, following this progressive extraction, TSS simultaneously calibrating the model towards the clean and informative data regime~\citep{wang2021survey,bengio2009curriculum}. Our theoretical analysis further substantiates this claim. In a nutshell, our contributions can be summarized as follows:

\begin{itemize}[leftmargin=14pt,topsep=1pt, itemsep=3pt]
    \item We identify the challenge of the previous sample selection with the noisily labeled graph data, and propose a TSS method to address the dilemma, which considers the topological characteristic to effectively select clean informative nodes for training under label noise.
    \item The proposed TSS method progressively extracts informative nodes and reduces the negative impact of label noise. We provide theoretical proof demonstrating that our method consistently minimizes an upper bound of the expected risk under the target clean distribution. 
    \item We conduct extensive experiments on various benchmarks to show the superiority of the proposed method over the state-of-the-art baselines in learning with noisily labeled graph, and provide comprehensive verification about the underlying mechanism of our method.
\end{itemize}

\section{Method}

\subsection{Notations and Preliminary}

Assume that we have an undirected graph $\mathcal{G} = (\mathcal{V}, \mathcal{E})$, where $\mathcal{V}=\{\mathbf{v}_1,...,\mathbf{v}_n\}$ is the set of $n$ nodes, $\mathcal{E} \subseteq \mathcal{V} \times \mathcal{V}$ is the set of edges, and $\mathbf{A} \in \mathbb{R}^{n \times n}$ is the adjacency matrix of the graph $\mathcal{G}$. If nodes $\mathbf{v}_i$ and $\mathbf{v}_j$ are connected by edges  $(\mathbf{v}_i,\mathbf{v}_j) \in \mathcal{E}$, $\mathbf{A}_{ij} = 1$; otherwise, $\mathbf{A}_{ij} = 0$. Let $\mathbf{D} \in \mathbb{R}^{n \times n}$ be the diagonal matrix, and $\hat{\mathbf{A}} \in \mathbb{R}^{n \times n} $ be the normalized adjacency matrix $\mathbf{D}^{-1/2}\mathbf{A}\mathbf{D}^{-1/2}$. Denote $\mathcal{X} = \{\mathbf{x}_1,...,\mathbf{x}_n\}$ and $\mathcal{Y}= \{y_1,...,y_n\}$ as the sets of node attributes and node labels respectively, with ${\mathbf{x}}_i$ being the node attribute of node $\mathbf{v}_i$ and $y_i$ being the true label of node $\mathbf{v}_i$. In this study, we have a dataset $\mathcal{D}=\{\tilde{\mathcal{D}}_{\mathrm{tr}}, \mathcal{D}_{\mathrm{te}}\}$, where  $\tilde{\mathcal{D}}_{\mathrm{tr}}=\{(\mathbf{A},\mathbf{x}_i,\tilde{y}_i)\}^{n_{\mathrm{tr}}}_{i=1}$ is a noisy training set drawn from a noisy distribution $\mathbbm{P}_{\tilde{\mathcal{D}}}=\mathbbm{P}(\mathbf{A},\mathcal{X},\tilde{\mathcal{Y}})$ ($\tilde{y}_i$ is the noisy counterpart of $y_i$), and $\mathcal{D}_{\mathrm{te}}$ is a clean test set drawn from a clean distribution $\mathbbm{P}_{\mathcal{D}}=\mathbbm{P}(\mathbf{A},\mathcal{X},\mathcal{Y})$. Our goal is to learn a proper GNN classifier $f_{\mathcal{G}}: (\mathbf{A}, \mathcal{X}) \rightarrow \mathcal{Y}$ from the noisy training set $\tilde{\mathcal{D}}_{\mathrm{tr}}$.

\subsection{Class-conditional Betweenness Centrality}

As discussed in the introduction, when extracting clean nodes from a noisily labeled graph, the roles of nodes vary greatly based on their topological information~\citep{wei2023clnode,song2022tam}. Some nodes situated near topological class boundaries are more difficult to be extracted since they lack the typical characteristics of their corresponding classes when aggregating neighbour nodes of different classes~\citep{wei2023clnode}. To alleviate this issue, we introduce a \textit{Class-conditional Betweenness Centrality} (CBC) measure that takes into account the topological structure of nodes, formulated as follows.

\begin{definition}[Class-conditional Betweenness Centrality] \label{Robust Class-conditional Betweenness Centrality}
Given the Personalized PageRank matrix $\boldsymbol{\pi} = \alpha(\mathbf{I}-(1-\alpha)\hat{\mathbf{A}})^{-1}$ ($\boldsymbol{\pi} \in \mathbb{R}^{n \times n}$), the Class-conditional Betweenness Centrality of the node $\mathbf{v}_i$ is defined by counting how often the node $\mathbf{v}_i$ is traversed by a random walk between pairs of other nodes that belong to different classes in a graph $\mathcal{G}$:
\begin{equation} \label{RCBC_equation_im}
\mathbf{Cb}_{i} \vcentcolon=  \frac{1}{n(n-1)}\sum_{\substack{\mathbf{v}_u \neq\mathbf{v}_i \neq\mathbf{v}_v, \\ \tilde{y}_u \neq \tilde{y}_v}}
\frac{\boldsymbol{\pi}_{u,i}\boldsymbol{\pi}_{i,v}}{\boldsymbol{\pi}_{u,v}},
\end{equation}
where $\boldsymbol{\pi}_{u,i}$ with the target node $\mathbf{v}_u$ and the source node $\mathbf{v}_i$ denotes the probability that an $\alpha$-discounted random walk from node $\mathbf{v}_u$ terminates at $\mathbf{v}_i$. An $\alpha$-discounted random walk represents a random traversal that, at each step, either terminates at the current node with probability $\alpha$, or moves to a random out-neighbour with probability $1 - \alpha$.
\end{definition}

\begin{figure*} [t]
\centering
\includegraphics[width=1\linewidth]{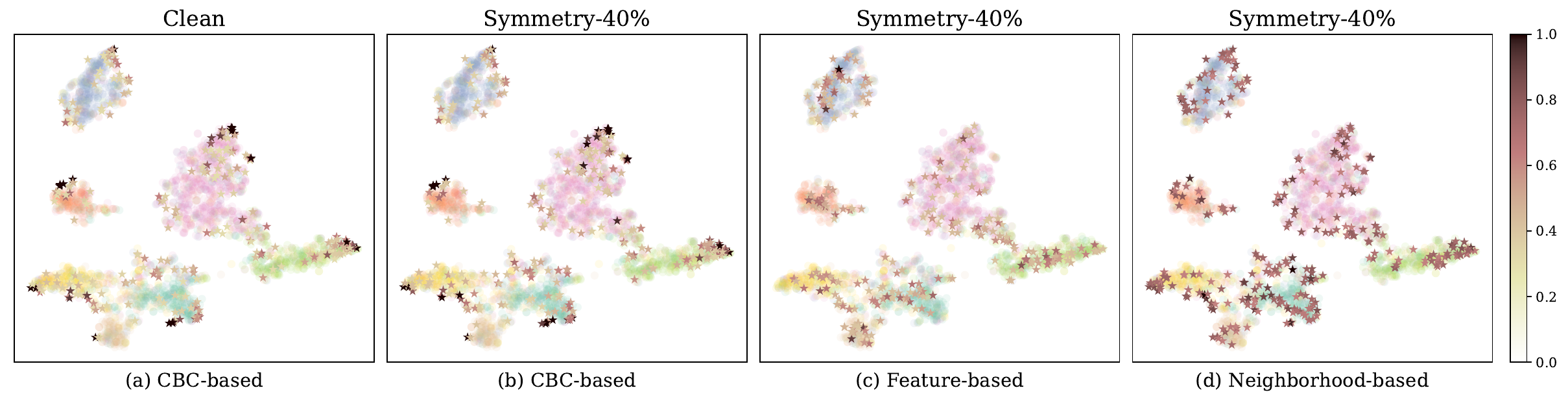}
\vspace{-15pt}
\caption{Robustness of Class-conditional Betweenness Centrality (\textit{t-SNE} visualization of node embeddings based on trained GNNs from the \textit{CORA} dataset). \textbf{(a)} clean labeled nodes with less CBC (lighter colour) are farther-away from class
boundaries than those with high CBC (darker colour). \textbf{(b)(c)(d)} Compared with other two difficulty measurers~\citep{wei2023clnode,li2023curriculum} in graph curriculum learning under 40\% \textit{Symmetric} label noise, CBC clearly shows superiority in terms of the differentiation \textit{w.r.t.} boundary-near nodes.}
\label{figure::CBC_effective}
\vspace{-10pt}
\end{figure*}

Note that, CBC is inspired by the classical concept in graph theory -- \textit{Betweenness Centrality}~\citep{newman2005measure,brandes2001faster} that measures the centrality of nodes in a graph~\footnote{In graph theory, the betweenness of a node $\mathbf{v}_i$ is defined to be the fraction of shortest paths between pairs of nodes in a graph that passes through $\mathbf{v}_i$. We provide its formal definition and discussion in Appendix.}, but significantly differs from the class-conditional constraint and the random walk realization instead of the short-path counting. We kindly refer the readers to the Appendix~\ref{CBC_diss} for the detailed discussion about their difference.

\begin{figure*} [t]
\centering
\includegraphics[width=0.9\linewidth]{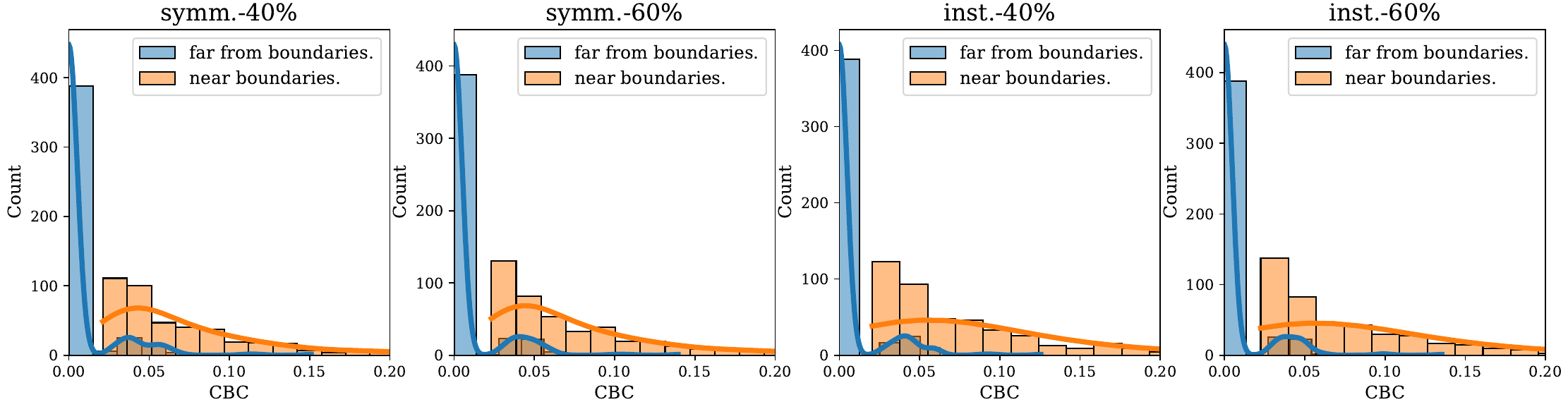}
\caption{The distributions of the CBC score \textit{w.r.t.} nodes on WikiCS with $40\%$ and $60\%$ symmetric noise (symm.) or $40\%$ and $60\%$ instance-based noise (inst.). The nodes are considered ``far from topological class boundaries" (far from boundaries.) when their two-hop neighbours belong to the same class; conversely, nodes are categorized as ``near topological class boundaries" (near boundaries.) when this condition does not hold. More comprehensive experiments in the Appendix~\ref{CBC_diss}.}
\label{figure::CBC_distribution}
\vspace{-10pt}
\end{figure*}

\vspace{-10pt}
\paragraph{Robustness of Class-conditional Betweenness Centrality} 
One promising merit of CBC is that it is robust to the label noise, although by definition it is based on the pair of nodes from different classes. As shown in Fig.~\ref{figure::CBC_effective} (b), under the high rate of label noise, the CBC of each node still can be accurately measured and the performance is close to the Fig.~\ref{figure::CBC_effective} (a) under clean labels. We also compare the performance of CBC with the other two difficulty measurers~\citep{li2023curriculum} in the Fig.~\ref{figure::CBC_effective} (c) and (d) to demonstrate our effectiveness. This is because CBC just requires that the node pairs belong to different classes instead of their absolutely accurate class labels, which is compatible with the general noise-agnostic scenarios. For example, if we have a pair of nodes whose latent true labels $(y_1 = 1, y_2 = 2)$ corresponding to the obvious noisy labels $(\tilde{y}_1 = 1, \tilde{y}_2 = 3)$, this node pair would not hinder the computation of CBC. Besides, even if the node pair actually belongs to the same underlying true class, CBC then degrades to the Betweenness Centrality and does not heavily hurt the total measure. 
Additionally, to demonstrate the consistent robustness of our CBC under varying levels of label noise, we visualize the superiority of our CBC distribution with numerical results as Fig.~\ref{figure::CBC_distribution}. The node dataset exhibits two distinct clusters, and despite a significant extent of label noise, certain nodes located near topological class boundaries consistently receive higher CBC scores. The complete and related experiment details have been presented in Appendix~\ref{CBC_diss}.

\begin{figure} [h]
\centering
\includegraphics[width=1\linewidth]{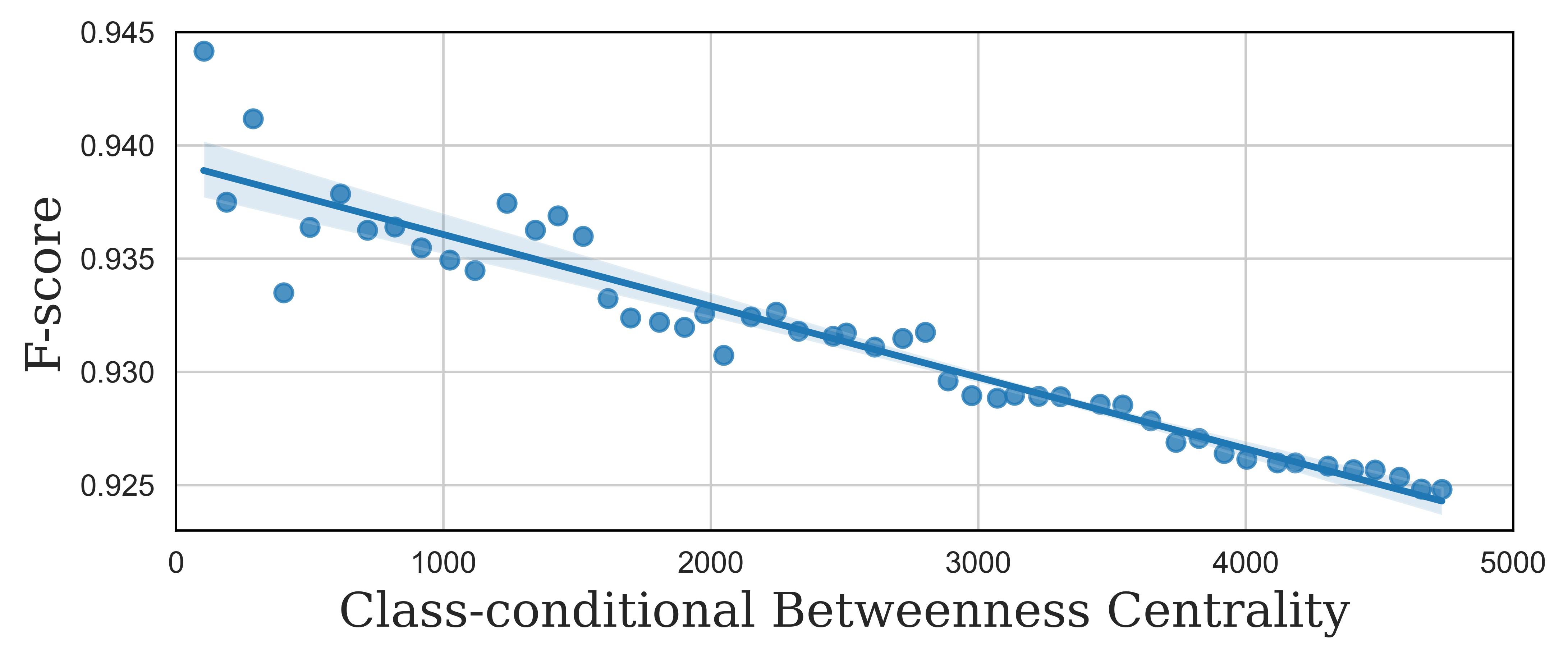}
 \caption{Correlation between \textit{F-score} of extracting confident nodes and overall CBC of the noisily labeled subsets in a graph with 30\% \textit{Symmetric} label noise. The Pearson coefficient is $-0.9276$ on 50 randomly selected subsets with $p$ value smaller than $0.0001$.}
\label{cbc_difficult_selection}
\vspace{-15pt}
\end{figure}

\vspace{-10pt}
\paragraph{Effectiveness of Class-conditional Betweenness Centrality}

To demonstrate the effectiveness of~Eq. (\ref{RCBC_equation_im}), we conduct an empirical verification presented in Fig.~\ref{cbc_difficult_selection}. As observed, the ability to extract clean nodes from the subset of noisily labeled nodes notably diminishes as CBC increases, consistent with the expected behaviour of CBC. Additionally, nodes with elevated CBC values tend to be situated closer to the decision boundary, which is essential to characterize the decision boundary for classifier~\citep{bengio2009curriculum,he2018decision,huang2010active,vapnik1999nature,bai2021me}. Leveraging the CBC measure allows us to selectively choose more informative nodes, significantly enhancing GNNs' performance during the training process. For further empirical evidence demonstrating the positive correlation between test accuracy and the overall CBC of the training set, we kindly refer readers to Appendix~\ref{CBC_diss}.

\subsection{Topological Sample Selection}

In this section, we construct the Topological Sample Selection (TSS) method leveraging the CBC measure to enhance informative sample selection in the presence of label noise. Utilizing the CBC measure, the TSS process begins by extracting clean nodes situated far from class boundaries and fitting a model on them. Then the model in TSS can learn clean patterns from the fitted clean nodes, which makes it possible for TSS to extract clean nodes from those nodes near class boundaries and entangled with incorrect labels. A visualization is shown in the Appendix~\ref{app:E}. Specifically, this process is similar to curriculum learning~\citep{bengio2009curriculum,guo2018curriculumnet} as it also learns from easy ones (extracted nodes located far from class boundaries) to hard (extracted nodes close to class boundaries) ones, with their identification guided by the CBC measure. Next, we formula TSS from the perspective of curriculum learning.

Here, we devise an ``easy-to-hard" curriculum within our TSS method, building upon the CBC. This curriculum is structured as a sequence of training criteria $\langle \tilde{Q}_{\lambda}\rangle$ with the increasing pace parameter $0 \leq \lambda \leq 1$. Each criterion $\tilde{Q}_{\lambda}$ is a reweighting of the noisy training distribution $\mathbbm{P}_{\tilde{\mathcal{D}}}$. The early $\tilde{Q}_{\lambda}$ emphasises the easy nodes (located far from class boundaries) evaluated by CBC, and as $\lambda$ increases, more hard nodes (closer to class boundaries) are progressively added into $\tilde{Q}_{\lambda}$, detailed in the following. Note that, while several methods may involve in curriculum learning, few of them address noisy labeled graphs by considering the intricate graph structure~\footnote{More discussion of related works has been summarized in the Appendix~\ref{related_work} due to the space limitation.}.

\vspace{-10pt}
\paragraph{Extracting Clean Labeled Nodes} The extraction of clean labeled nodes is closely related to the \textit{memorization effect} of neural networks~\citep{arpit2017closer,lin2023over,xia2020robust}. Specifically, due to the memorization effect, the GNN classifier trained at early epochs would fit the clean data well but not the incorrectly labeled data. We can treat the training nodes whose noisy labels are identical to the ones predicted by the trained classifier as the confident nodes, indicating a higher likelihood of having clean labels~\citep{bai2021me}. Note that, there are other similar rules to extract clean examples, e.g., those who have a high confidence score or corresponding to a smaller loss value~\citep{han2018co,yu2019does,xia2021sample,li2024instant}, which will be compared in experiments. Now, we progressively obtain the extracted nodes, which are more likely to have clean labels and named as the confident nodes~\citep{bai2021me,xia2021sample}. The extracted confident node set $\mathbbm{P}_{\hat{\mathcal{D}}}$ from  $\mathbbm{P}_{\tilde{\mathcal{D}}}$, which approximates nodes drawn from a target clean distribution $\mathbbm{P}_{{\mathcal{D}}}$. With $\mathbbm{P}_{\hat{\mathcal{D}}}$, we can construct our robust learning curriculum.
\vspace{5pt}

\begin{definition}[Topological Sample Selection] \label{Topological Curriculum Learning}
Assume a sequence of extracted confident training criteria $\langle \hat{Q}_{\lambda}\rangle$ with the increasing pace parameter $\lambda$. Each confident criterion $\hat{Q}_{\lambda}$ is a reweighting of the confident distribution $\mathbbm{P}_{\hat{\mathcal{D}}}(z)$, where $z$ is a random variable representing an extracted confident node for the learner. Let $0 \leq W_{\lambda}(z) \leq 1 $ be the weight on $z$ at step $\lambda$ in the curriculum sequence, and 
\begin{equation} \label{TSS}
\hat{Q}_{\lambda}(z) \propto W_{\lambda}(z) \mathbbm{P}_{\hat{\mathcal{D}}}(z),
\end{equation}
such that $\int_{\mathcal{Z}}\hat{Q}_{\lambda}(z)dz = 1$, where $\mathcal{Z}$ denotes the whole set of extracted confident nodes from each $\hat{Q}_{\lambda}(z)$. Then, the following two conditions are satisfied:
\vspace{-5pt}
\begin{itemize}
    \item (i) The entropy of distributions gradually increases, i.e., $H(\hat{Q}_{\lambda})$ is monotonically increasing with respect to $\lambda$.
    \item (ii) The weight function $W_{\lambda}(z)$ for any confident nodes is monotonically increasing with respect to $\lambda$.
\end{itemize}
\end{definition}

In Definition~\ref{Topological Curriculum Learning}, Condition (i) means that the diversity and the information of the extracted confident set should gradually increase, \ie the reweighting of nodes in later steps increases the probability of sampling informative nodes evaluated by CBC. Condition (ii) means that as gradually adding more confident nodes, the size of the confident node set progressively increases. Intuitively, in our TSS, the key is the proposed CBC measure that works as a difficulty measurer and defines the weight function $W_{\lambda}(z)$. This formalization has been widely used in the related curriculum learning literature~\citep{bengio2009curriculum,wang2021survey}. With the help of CBC, we can design a robust ``easy-to-hard" learning curriculum that first extracts confident nodes from noisily easy nodes (located far away from class boundaries) – that we term as \textit{high-confident nodes} to train GNNs and then extracts confident nodes from noisily hard nodes (close to class boundaries) – that we term as \textit{low-confident nodes} to continually train. We summarize the procedure of TSS in Algorithm~\ref{alg:TSS} of the Appendix.

\vspace{-5pt}
\subsection{Theoretical Guarantee of TSS}

Here, we first investigate the change in deviation between  ${\mathbbm{P}}_{\hat{\mathcal{D}}}$ and ${\mathbbm{P}}_{\mathcal{D}}$ during the learning phases of TSS. Then, we theoretically prove that, with the ${\mathbbm{P}}_{\hat{\mathcal{D}}}$, our TSS method persistently minimizes an upper bound of the expected risk under target clean distribution. 

Taking the binary node classification as an example, after extracting the confident nodes, our goal is to learn a proper GNN classifier $f_{\mathcal{G}}: (\mathbf{A}, \mathcal{X}) \rightarrow \mathcal{Y}$ with the input extracted confident nodes $z_{i} =\{(\mathbf{A},\mathbf{x}_i,y_i)\}^{n_{\mathrm{cf}}}_{i=1}$ from the confident distribution $\mathbbm{P}_{\hat{\mathcal{D}}}(\mathcal{Z}) = \mathbbm{P}_{\hat{\mathcal{D}}}(\mathbf{A},\mathcal{X}|\mathcal{Y})\mathbbm{P}_{\hat{\mathcal{D}}}(\mathcal{Y})$~\citep{cucker2007learning}, such that the following expected risk can be minimized:
\begin{equation}
\mathcal{R}(f_{\mathcal{G}}):= \int_{Z}\mathcal{L}_{f_{\mathcal{G}}}(z)\mathbbm{P}_{\mathcal{D}}(\mathbf{A},\mathbf{x}|y)\mathbbm{P}_{\mathcal{D}}(y)dz,
\end{equation}
where $\mathbbm{P}_{{\mathcal{D}}}(\mathcal{Z}) = \mathbbm{P}_{{\mathcal{D}}}(\mathbf{A},\mathcal{X}|\mathcal{Y})\mathbbm{P}_{{\mathcal{D}}}(\mathcal{Y})$ denotes the target clean distribution on $\mathcal{Z}$, and $\mathcal{L}_{f_{\mathcal{G}}}(z)= \mathbbm{1}_{f_{\mathcal{G}}(\mathbf{A},\mathbf{x}) \neq y} = \frac{1-yf_{\mathcal{G}}(\mathbf{A},\mathbf{x})}{2}$ denotes the loss function measuring the difference between the predictions and labels. Since the deduction for both $y = 1$ and $y = -1$ cases are exactly similar, we only consider one case in the following and denote $\mathbbm{P}_{{\mathcal{D}}}(\mathbf{A}, \mathbf{x}) = \mathbbm{P}_{{\mathcal{D}}}(\mathbf{A}, \mathbf{x}|y=1)$ and $\mathbbm{P}_{{\mathcal{\hat{D}}}}(\mathbf{A}, \mathbf{x}) = \mathbbm{P}_{{\mathcal{\hat{D}}}}(\mathbf{A}, \mathbf{x}|y=1)$.
Let $0 \leq W_{\lambda^{*}}(\mathbf{A}, \mathbf{x}) \leq 1$, $\alpha^{*} = \int_{\mathbf{A},\mathcal{X}}W_{\lambda^{*}}(\mathbf{A}, \mathbf{x})\mathbbm{P}_{{\hat{\mathcal{D}}}}(\mathbf{A}, \mathbf{x})d\mathbf{x}$ denote the normalization factor\footnote{The $\alpha^{*} \leq 1$ since $W_{\lambda^{*}}(\mathbf{A}, \mathbf{x})\mathbbm{P}_{{\hat{D}}}(\mathbf{A}, \mathbf{x}) \leq \mathbbm{P}_{{\hat{D}}}(\mathbf{A}, \mathbf{x}) $} and $E(\mathbf{A}, \mathbf{x})$ measures the deviation from $\mathbbm{P}_{{\hat{\mathcal{D}}}}(\mathbf{A}, \mathbf{x})$. Combining with Definition~\ref{Topological Curriculum Learning}, we can construct the below curriculum sequence for theoretical evaluation (See proof in the Appendix~\ref{theoretical_proof}):
\begin{equation} \label{theoretical cl}
\hat{Q}_{\lambda}(\mathbf{A},\mathbf{x}) \propto W_{\lambda}(\mathbf{A},\mathbf{x}) \mathbbm{P}_{\hat{\mathcal{D}}}(\mathbf{A},\mathbf{x}), 
\end{equation}
where 
\begin{equation}
W_{\lambda}(\mathbf{A},\mathbf{x}) \propto \frac{\alpha_{\lambda}\mathbbm{P}_{{\mathcal{D}}}(\mathbf{A}, \mathbf{x}) + (1-\alpha_{\lambda})E(\mathbf{A},\mathbf{x})}
{\alpha^{*}\mathbbm{P}_{{\mathcal{D}}}(\mathbf{A}, \mathbf{x})+(1-\alpha^{*})E(\mathbf{A},\mathbf{x})} \nonumber
\end{equation}
with $0 \leq W_{\lambda}(\mathbf{A}, \mathbf{x}) \leq 1$ through normalizing its maximal value as 1 and $\alpha_{\lambda}$ varies from 1 to $\alpha^{*}$ with increasing pace parameter $\lambda$. Note that, the initial stage of TSS sets $W_{\lambda}(\mathbf{A},\mathbf{x}) \propto \frac{\mathbbm{P}_{{\mathcal{D}}}(\mathbf{A}, \mathbf{x})}{\mathbbm{P}_{{\hat{D}}}(\mathbf{A}, \mathbf{x})}$, which is of larger
weights in the high-confident nodes while much smaller in low-confident nodes. With the pace $\lambda$ increasing, the large weights in high-confidence areas become smaller while small ones in low-confidence areas become larger, leading to more uniform  weights with smaller variations. 

Here, we introduce a \textit{local-dependence} assumption for graph-structured data: Given
the data related to the neighbours within a certain number of hops of a node $\mathbf{v}_i$, the data in the rest of the graph will be independent of $\mathbf{v}_i$~\citep{wu2020graph}. This assumption aligns with Markov chain principles~\citep{revuz2008markov}, stating that the node is independent of the nodes that are not included in their two-hop neighbors when utilizing two-layer GNN, which does not means the totally i.i.d w.r.t. each node but means i.i.d w.r.t. subgroups. The local-dependence assumption is well-established and has been widely adopted in numerous graph theory studies~\citep{schweinberger2015local,didelez2008graphical}. It endows models with desirable properties which make them amenable to statistical inference~\citep{schweinberger2015local}. Therefore, based on the local-dependence assumption, for a node with the certain hops of neighbours $Z^\mathbf{A}$, after aggregation, we will obtain node representation $Z_{\mathbf{x}_i}$ that is approximately independent and identically distributed with nodes outside of $Z^\mathbf{A}$. We  refer readers to ~\citep{gong2016curriculum} for more details. Finally, with Eq. (\ref{theoretical cl}) as the pace distribution, we have the following theorem and a detailed proof is provided in Appendix~\ref{theoretical_proof}.

\begin{theorem}
\label{bound}
Suppose $\{(Z_{\mathbf{x}_i},y_i)\}^{m}_{i=1}$ are i.i.d. samples drawn from the pace distribution $Q_{\lambda}$ with radius $|X| \leq R$. Denote $m_{+}/m_{-}$ be the number of positive/negative samples and $m^{*} = \min\{m_{-},m_{+}\}$. Let $\mathcal{H} = \{\mathbf{x} \rightarrow \mathbf{w}^{T}\mathbf{x}: min_{s}|\mathbf{w}^{T}\mathbf{x}| = 1 \cap ||\mathbf{w}|| \leq B \}$, and $\phi(t) = (1-t)_{+}$ for $t \in \mathbbm{R}$ be the hinge loss function. For any $\delta > 0$ and $g \in \mathcal{H}$, with confidence at least $1-2\delta$, have:
\begin{equation}
\small
    \begin{split}
        \mathcal{R}(\text{sgn}(g)) &\leq \frac{1}{2m_{+}}\sum_{i=1}^{m_{+}}\phi(y_ig(Z_{\mathbf{x}_i}))+\frac{1}{2m_{-}}\sum_{i=1}^{m_{-}}\phi(y_ig(Z_{\mathbf{x}_i}))\\ 
        &+\frac{RB}{\sqrt{m^{*}}} + 3\sqrt{\frac{\ln{(1-\delta)}}{m^{*}}}\\
        &+(1-\alpha_{\lambda})\sqrt{1-\exp{\{-D_{KL}(\mathbbm{P}^{+}_{\mathcal{D}}||E^{+})\}}}\\
        &+(1-\alpha_{\lambda})\sqrt{1-\exp{\{-D_{KL}(\mathbbm{P}^{-}_{\mathcal{D}}||E^{-})\}}},\\
        &
    \end{split}
\end{equation}
where $E^{+}$, $E^{-}$ denote error distributions that capture the deviation from $\mathbbm{P}^{+}_{\mathcal{D}}$, $\mathbbm{P}^{-}_{\mathcal{D}}$ to $\mathbbm{P}^{+}_{\mathcal{\hat{D}}}$, $\mathbbm{P}^{-}_{\mathcal{\hat{D}}}$.
\end{theorem}

\begin{figure*}[t]
\centering
\begin{subfigure}{0.45\linewidth}
  \centering
  \includegraphics[width=\linewidth,height=2.7cm]{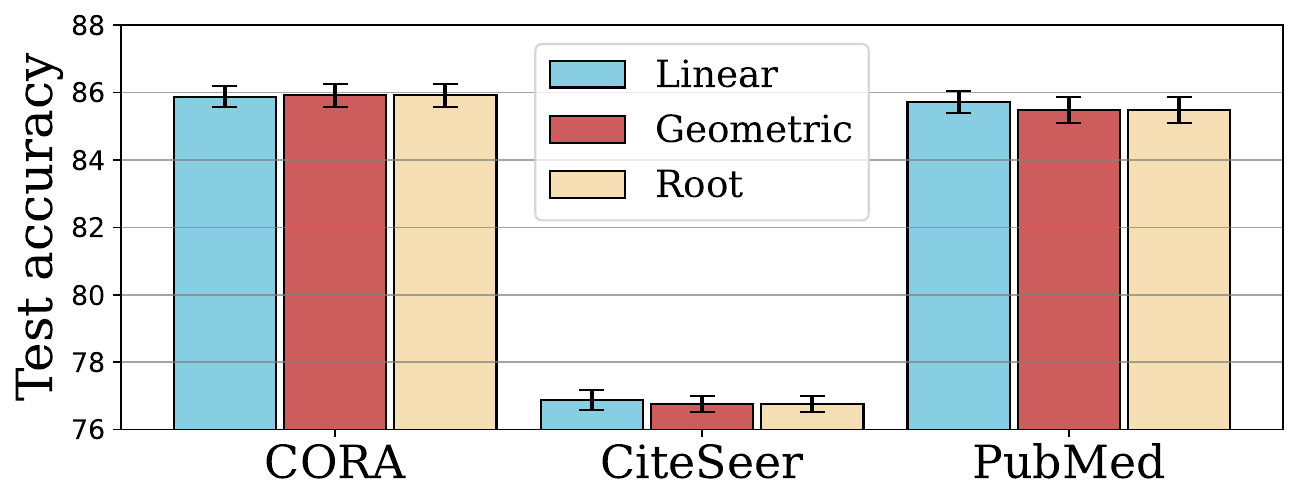}
\end{subfigure}
 \hspace{0.1in}
\begin{subfigure}{0.45\linewidth}
  \centering
  \includegraphics[width=\linewidth,height=2.7cm]{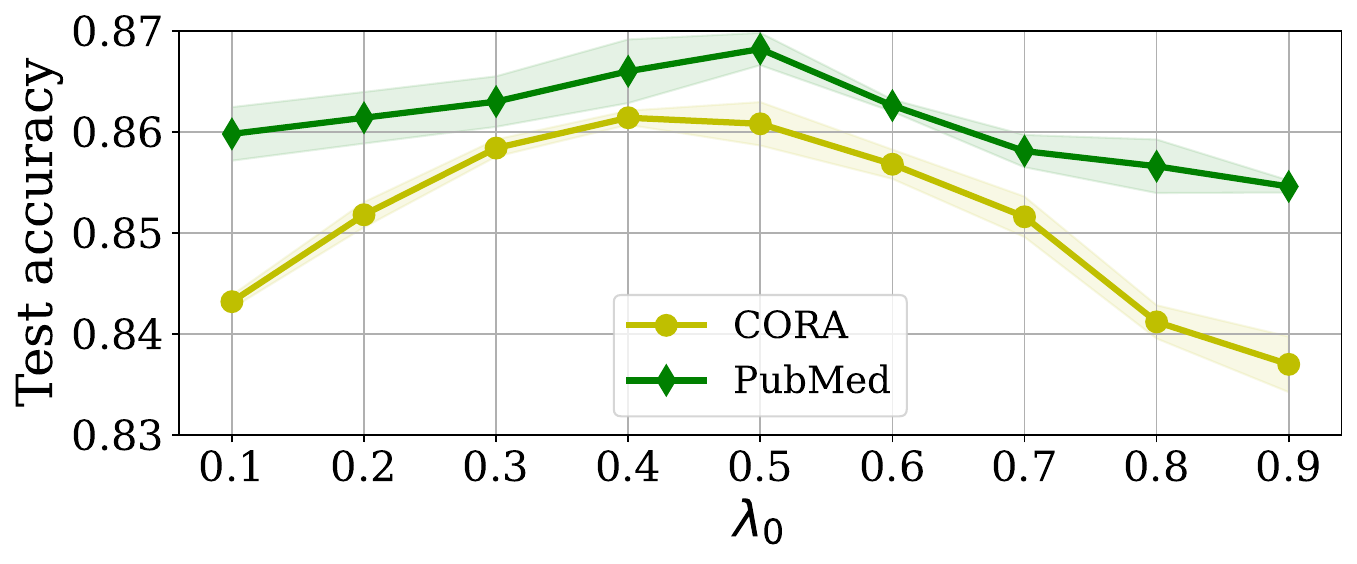}
\end{subfigure}
\vspace{-5pt}
  \caption{The hyperparameter analysis of TSS. The experiment results are reported over five trials under the 20\% \textit{Symmetric} noise. \textbf{(a)} The test accuracy of TSS with three different pacing functions on various datasets. \textbf{(b)} The test accuracy of TSS with increasing $\lambda_{0}$ on \textit{CORA} and \textit{PubMed}.}
\label{figure::hyper}
\vspace{-15pt}
\end{figure*}

\paragraph{Remark 1} (on the upper bound of the expected risk $\mathcal{R}(\text{sgn}(g)$). The error distribution $E$ reflects the difference between the noisy distribution and the clean distribution. Essentially, this error distribution serves as a bridge connecting the noisy and clean distributions in our upper bound. Thus, the last two rows measure the generalization capability of the learned classifier, which is monotonically increasing with respect to both the KL-divergence between the error distribution $E$ and the clean distribution $\mathbbm{P}_{\mathcal{D}}$, and the pace parameter $\lambda$. That is, the less deviated is the error $E$ from $\mathbbm{P}_{\mathcal{D}}$, the more beneficial is to learn a proper classifier from $\mathbbm{P}_{\hat{\mathcal{D}}}$ which can generalize well on  $\mathbbm{P}_{\mathcal{D}}$. 

Thus, the TSS process with curriculum $\hat{Q}_{\lambda}$ makes it feasible to approximately learn a graph model with minimal expected risk on $\mathbbm{P}_{{\mathcal{D}}}$ through the empirical risk from $\mathbbm{P}_{\hat{\mathcal{D}}}$, since the "easy-to-hard" property of the curriculum $\hat{Q}_{\lambda}$ intrinsically facilitates the information transfer from $\mathbbm{P}_{\hat{\mathcal{D}}}$ to $\mathbbm{P}_{{\mathcal{D}}}$. In specific, we can approach the task of minimizing the expected risk on $\mathbbm{P}_{{\mathcal{D}}}$ by gradually increasing the pace $\lambda$, generating relatively high-confident nodes from $\hat{Q}_{\lambda}$, and minimizing the empirical risk on those nodes. This complies with the core idea of the proposed TSS. In addition, the first row in the upper bound of Theorem~\ref{bound} corresponds to the empirical risk on training nodes generated from $Q_{\lambda}$. The second row reflects that the more training nodes are considered, the better approximation of expected risk can be achieved~\citep{haussler1993probably,haussler1990probably}.

\section{Experiments}

In this section, we conduct extensive experiments to verify the effectiveness of our method and provide  comprehensive ablation studies about the underlying mechanism of TSS.

\paragraph{Datasets} We adopted three small datasets including \textit{Cora}, \textit{CiteSeer}, and \textit{PubMed}, with the default dataset split as did in~\citep{chen2018fastgcn}, and four large datasets: \textit{WikiCS}, \textit{Facebook}, \textit{Physics} and \textit{DBLP} to evaluate our method. Detailed statistics are summarized in Appendix. Following previous works~\citep{dai2021nrgnn,du2021pi,xia2020part}, we consider three settings of simulated noisy labels, i.e, \textit{Symmetric} noise, \textit{Pairflip} noise and \textit{Instance-dependent} noise. More explanation about noise settings in Appendix~\ref{lng}.
\vspace{-10pt}

\paragraph{Baselines} 
We compare TSS with several state-of-the-art sample selection with noisy labels on i.i.d. data: (1) Co-teaching+~\citep{yu2019does}, (2) Me-Momentum~\citep{bai2021me} and (3) MentorNet~\citep{yu2019does}. we also compare with the graph curriculum learning method: (1) CLNode~\citep{wei2023clnode}, (2) RCL~\citep{zhang2023relational}. Besides, some denoising methods on graph data have been considered (1) LPM~\citep{xia2020towards}, (2) CP~\citep{zhang2020adversarial}, (3) NRGNN~\citep{dai2021nrgnn}, (4) PI-GNN~\citep{du2023noise}, (5) RT-GNN~\citep{qian2023robust} and (6)RS-GNN~\citep{dai2022towards}. More details about implementations are provided in the Appendix~\ref{baseline}.

\subsection{Main results}

\begin{table*}[t]
	\caption{Mean and standard deviations of classification accuracy (percentage) on synthetic noisy
datasets with different noise levels. The results are reported over ten trials and the best are bolded.}
\vspace{-5pt}
	\centering
	 \small
	\setlength\tabcolsep{3pt}

	\resizebox{1\linewidth }{!}{
         \midsepremove
         
	\begin{tabular}{l|c| ccccccccc}
		\toprule
		 & \multirow{2}{*}{Method} & \multicolumn{3}{c}{\textit{Symmetric}} & \multicolumn{3}{c}{\textit{Pairflip}} & \multicolumn{3}{c}{\textit{Instance-dependent}}\\
            \cline{3-11}
		& & 30\% & 40 \%& 50\% &  20\% & 30\% & 40\% & 30\% & 40\% & 50\%\\
		
  \midrule	
		\multirow{12}{*}{\rotatebox{90}{\textit{CORA}}}
    & Cross-Entropy & {83.61±1.07}& 80.86±1.46 & {75.14±2.44} &{82.23±0.93} & {75.87±1.20} &{62.05±3.59} &{83.21±0.74}&{80.32±0.94} &{74.96±1.82}\\
    
   & LPM &\makecell[c]{82.73±0.64} &78.12±1.17& \makecell[c]{70.23±2.17} & \makecell[c]{83.39±1.22}&77.44±1.93&\makecell[c]{64.02±5.04} &\makecell[c]{82.81±0.87} &77.67±2.01& \makecell[c]{70.55±1.86}\\
    
    & CP & \makecell[c]{82.37±1.38} &79.97±1.74& \makecell[c]{76.19±2.26} & \makecell[c]{80.24±0.96}&73.02±1.56& \makecell[c]{58.04±3.78} &\makecell[c]{82.37±1.09} &80.36±1.21& \makecell[c]{74.17±2.68}\\

		& NRGNN &\makecell[c]{81.73±1.80} &79.08±3.18& \makecell[c]{77.36±2.03}& \makecell[c]{81.83±0.93}&77.10±1.52& \makecell[c]{64.13±3.98}&\makecell[c]{81.62±2.08} &78.66±2.54& \makecell[c]{76.31±2.98}\\
  
  & PI-GNN & 82.48±0.10 &	80.36±0.10 &	77.59±0.20 &	83.10±0.10	 &77.96±0.20	 &63.62±0.30 & 81.83±1.00 &80.02±1.07 & 77.27±1.21\\

    & RT-GNN & 83.21±1.05& 	80.46±1.06	& 75.84±1.43	& 82.53±0.73	& 76.87±1.09	& 61.75±2.29 & 82.14±0.97& 	80.13±1.23	& 74.82±0.94\\

  & RS-GNN & 83.21±0.29	& 79.00±0.15	& 77.21±0.43	& 81.83±0.37	& 76.46±0.24	& 63.09±0.22 & 82.83±0.73	& 78.93±0.63	& 76.52±0.83\\
  
\cline{2-11}	

		& Co-teaching+ & \makecell[c]{82.59±0.96} &79.81±1.30 & \makecell[c]{74.59±2.33} & \makecell[c]{81.70±1.45}&75.59±2.13&\makecell[c]{59.03±5.76}& \makecell[c]{81.84±1.10} & 79.70±1.34 &\makecell[c]{73.36±2.54} \\	

		& Me-momentum &\makecell[c]{83.76±0.25} &81.82±0.72 & \makecell[c]{79.48±0.63}& \makecell[c]{84.09±0.48}&78.04±1.03&\makecell[c]{64.07±1.03} &\makecell[c]{83.14±0.25} &82.04±0.57& \makecell[c]{77.33±0.82}\\	
  
        & MentorNet & 81.84±0.86 &78.52±2.01 & 73.82±2.83 & 80.83±1.88 & 72.56±3.42 & 59.78±4.59 &81.59±0.92 & 78.49±1.63 & 72.41±3.66\\
                & CLNode & 80.98±1.50 & 77.11±2.25 &74.39±2.41 & 83.43±0.89 & 73.89±1.97 & 55.38±2.80 & 81.12±2.43 & 75.11±2.93 & 68.44±4.88\\ 

                & RCL & 73.20±0.12 &  76.36±1.09 & 63.40±0.73 & 71.06±0.48 & 65.30±0.80 & 51.34±0.42 & 69.20±1.00 & 59.30±0.13 & 54.16±2.25\\
		\cline{2-11} &
      \cellcolor{LightGreen}
         TSS &  \cellcolor{LightGreen}\textbf{\makecell[c]{85.02±0.12}}& \cellcolor{LightGreen}\textbf{82.58±0.92} & \cellcolor{LightGreen}\textbf{\makecell[c]{81.16±0.80}}& \cellcolor{LightGreen}\textbf{\makecell[c]{85.26±0.30}}&\cellcolor{LightGreen}\textbf{78.50±0.72}&\cellcolor{LightGreen}\textbf{\makecell[c]{65.15±1.53}}  & \cellcolor{LightGreen}\textbf{\makecell[c]{84.70±0.04}}&\cellcolor{LightGreen}\textbf{83.31±0.21} &\cellcolor{LightGreen}\textbf{\makecell[c]{80.15±0.36}}\\
		\bottomrule
        \multirow{12}{*}{\rotatebox{90}{\textit{CiteSeer}}}
        
        & Cross-Entropy & \makecell[c]{75.13±0.70} & \makecell[c]{73.85±0.85} & \makecell[c]{70.74±1.86} &\makecell[c]{76.61±0.53} & \makecell[c]{73.87±1.08}&\makecell[c]{62.92±4.11} & \makecell[c]{74.83±1.04} &\makecell[c]{73.22±0.71} & \makecell[c]{69.42±2.07}\\
        
        	&LPM &\makecell[c]{73.19±1.07} &\makecell[c]{69.54±1.37} & \makecell[c]{61.22±2.08}& \makecell[c]{75.08±0.76}& \makecell[c]{69.91±1.31}&\makecell[c]{58.86±4.28} &\makecell[c]{73.55±0.79} &\makecell[c]{69.32±1.76} & \makecell[c]{61.90±1.73}\\	

		&CP &\makecell[c]{73.26±1.22} &\makecell[c]{70.99±1.88} & \makecell[c]{63.74±2.55}& \makecell[c]{74.36±1.21}& \makecell[c]{68.21±2.56}&\makecell[c]{56.56±6.50} &\makecell[c]{73.45±0.72} &\makecell[c]{69.90±1.64} & \makecell[c]{64.61±2.74}\\

		&NRGNN &\makecell[c]{75.41±1.04} &\makecell[c]{73.52±1.46} & \makecell[c]{70.98±2.47}& \makecell[c]{75.72±1.04}& \makecell[c]{74.13±1.38}&\makecell[c]{63.60±4.83} &\makecell[c]{75.33±0.91} &\makecell[c]{74.36±1.45} & \makecell[c]{71.61±1.76}\\

          & PI-GNN & 73.55±0.14	&71.05±0.21 &	68.02±0.20 &	73.06±0.13 &	69.91±0.32	&60.62±0.41 &74.28±0.78 & 70.66±1.51 & 67.81±1.99 \\

    & RT-GNN & 74.64±0.72	&73.66±0.58	&71.36±0.65	&73.32±0.68	&65.78±1.33 &	62.38±0.56 &  73.94±0.52	&72.86±0.48	&71.02±0.25\\

  & RS-GNN & 74.93±0.65 &	73.65±0.45	&70.54±1.26	 &76.31±0.33	&73.27±0.38	& 61.42±2.01 & 75.03±0.25 &	72.85±0.15	&70.14±1.06 \\
  
		\cline{2-11}	
        
        &Co-teaching+ &\makecell[c]{71.01±2.83} &\makecell[c]{68.12±2.38} & \makecell[c]{61.65±4.27}& \makecell[c]{72.09±1.21}& \makecell[c]{68.25±2.91}&\makecell[c]{56.64±5.46} &\makecell[c]{70.80±3.08} &\makecell[c]{67.46±2.55} & \makecell[c]{62.12±2.81}\\

		&Me-Momentum & \makecell[c]{75.40±0.26}&\makecell[c]{74.41±0.56} & \makecell[c]{70.51±0.79} &\makecell[c]{76.93±0.47} &\makecell[c]{74.07±1.06} &\makecell[c]{63.96±0.97} &\makecell[c]{75.27±0.25} &\makecell[c]{74.24±0.45} & \makecell[c]{71.18±0.45}\\

        & MentorNet & 69.61±3.42 &66.87±3.78 & 60.21±2.67 & 71.96±1.81 & 66.14±4.98 & 54.20±6.25 & 70.56±2.55 & 64.90±4.72 & 60.95±4.93\\
        
                  & CLNode & 68.73±2.07 & 64.26±3.18 & 56.07±3.61 &69.11±3.15 &61.62±3.33 & 53.32±4.29 & 69.91±1.88 & 66.22±2.65 & 60.37±3.10\\ 
                  
      & RCL & 60.90±0.12 & 54.50±2.53 & 46.58±1.44 &65.00±0.13 &56.68±0.27 & 51.14±1.58 & 63.70±0.53 & 54.70±1.97 & 46.62±0.59\\ 
                  
		\cline{2-11} &

  \cellcolor{LightGreen}
       TSS & \cellcolor{LightGreen}\textbf{75.86±0.31} & \cellcolor{LightGreen}\textbf{74.77±0.79} & \cellcolor{LightGreen}\textbf{71.81±0.74} &\cellcolor{LightGreen}\textbf{77.25±0.44} & \cellcolor{LightGreen}\textbf{74.91±0.90} &\cellcolor{LightGreen}\textbf{65.36±1.27} & \cellcolor{LightGreen}\textbf{76.61±0.17}& \cellcolor{LightGreen}\textbf{75.61±0.29} & \cellcolor{LightGreen}\textbf{74.03±0.26}\\	
		\toprule
        	\multirow{12}{*}{\rotatebox{90}{\textit{PubMed}}} 
         
         & Cross-Entropy & \makecell[c]{85.98±0.50} & \makecell[c]{84.80±0.83} & \makecell[c]{82.83±1.55} &\makecell[c]{85.31±0.38} &\makecell[c]{83.31±0.58} & \makecell[c]{76.12±2.04} &\makecell[c]{85.29±0.27} &\makecell[c]{84.10±0.74} & \makecell[c]{82.45±2.96}\\	

 		&LPM  &\makecell[c]{85.33±0.70} &\makecell[c]{84.33±0.79} & \makecell[c]{82.31±0.89} & \makecell[c]{ 85.90±0.57}&\makecell[c]{84.63±0.34} &\makecell[c]{78.94±0.79}&\makecell[c]{85.51±0.52} &\makecell[c]{84.90±0.53} & \makecell[c]{83.12±1.18}\\

		&CP  &\makecell[c]{86.12±0.63} &\makecell[c]{85.01±0.65} & \makecell[c]{82.33±1.51}& \makecell[c]{86.13±0.36}& \makecell[c]{84.87±0.46}&\makecell[c]{78.81±0.77} &\makecell[c]{85.66±0.60} &\makecell[c]{84.92±0.99} & \makecell[c]{81.18±1.95}\\
  
  		&NRGNN &\makecell[c]{86.19±0.44} &\makecell[c]{84.99±1.16} & \makecell[c]{83.02±1.44} & \makecell[c]{86.26±0.81}& \makecell[c]{83.79±1.28}& \makecell[c]{75.83±2.72}&\makecell[c]{85.45±0.52} &\makecell[c]{85.07±1.15} & \makecell[c]{83.47±1.02}\\

            & PI-GNN & 86.16±0.06	&85.35±0.11	 &83.12±0.13 &	86.01±0.12 &	84.09±0.21	& 78.35±0.23 & 86.13±0.29 & 85.09±0.40& 83.22±0.85 \\

    & RT-GNN & 84.73±0.05 &	84.70±0.35	& 79.39±0.25 &	82.90±0.03 &	80.80±0.10	& 79.90±0.12 & 83.09±0.43 &	81.60±0.15	& 80.81±0.32 \\

  & RS-GNN & 85.38±0.42	&84.34±0.38 &	82.37±0.35 &	85.24±0.24	& 83.12±0.47 &	75.24±1.27 & 85.16±0.32	&84.14±0.14 &	83.07±0.15\\
		\cline{2-11}

		&Co-teaching+ &\makecell[c]{86.14±0.58} &\makecell[c]{85.01±0.74} & \makecell[c]{82.74±2.12}& \makecell[c]{85.37±1.90}& \makecell[c]{84.45±0.75}&\makecell[c]{77.31±5.38} &\makecell[c]{85.83±0.54} &\makecell[c]{84.65±1.47} & \makecell[c]{81.42±2.89}\\

		&Me-Momentum  &\makecell[c]{86.05±0.18} &\makecell[c]{85.66±0.78} & \makecell[c]{82.42±0.41}& \makecell[c]{85.78±0.26}& \makecell[c]{85.43±0.35}&\makecell[c]{80.34±0.41} &\makecell[c]{85.87±0.27} &\makecell[c]{84.37±0.40} & \makecell[c]{83.53±0.14}\\

        & MentorNet & 85.43±0.81& 84.55±1.33 & 82.84±0.92 &86.64±0.59 &84.83±0.92 & 74.36±6.01 &85.14±1.12 & 84.13±1.75& 80.38±3.99 \\

        & CLNode & 86.03±0.37&85.34±0.45 & 83.06±0.37& 86.27±0.42  & 85.15±0.38 & 81.12±0.44 & 85.23±0.37 & 84.61±0.39 & 83.63±0.51\\

	      & RCL & 82.40±0.24 & 80.30±0.15 & 76.40±0.14 &82.70±0.23 &82.66±0.69 & 81.30±0.20 & 82.10±0.12 & 80.30±0.12 & 74.90±0.19\\ 

     \cline{2-11}& 
         \cellcolor{LightGreen}
        TSS & \cellcolor{LightGreen}\textbf{86.69±0.32} & \cellcolor{LightGreen}\textbf{86.23±0.37} & \cellcolor{LightGreen}\textbf{83.53±0.23} & \cellcolor{LightGreen}\textbf{87.05±0.28} & \cellcolor{LightGreen}\textbf{86.30±0.22}& \cellcolor{LightGreen}\textbf{83.18±0.55} & \cellcolor{LightGreen}\textbf{86.21±0.03} & \cellcolor{LightGreen}\textbf{85.32±0.04} & \cellcolor{LightGreen}\textbf{83.94±0.08} \\	
		\bottomrule
	\end{tabular}
}
	\label{tab:classification}
 \vspace{-5pt}
\end{table*}

\paragraph{Performance comparison on public graph datasets}
Table~\ref{tab:classification} shows the experimental results on three synthetic noisy datasets under various types of noisy labels. For three datasets, as can be seen, our proposed method produces the best results in all cases. When the noise rate is high, the proposed method still achieves competitive results through the extraction of confident nodes. Although some baselines, \eg NRGNN, can work well in some cases, experimental results show that they cannot handle various noise types. In contrast, the proposed TSS achieves superior robustness against broad noise types. Lastly, some popular sample-selection methods that have worked well on learning with noisy labels on i.i.d. data, \eg Co-teaching+, do not show superior performance on graph data. This illustrates that the unique topology consideration in GNNs brings new challenges to those prior works and proves the necessity of TSS.

\paragraph{Performance comparison on large graph datasets}
We justify that the proposed methods can effectively alleviate the label noise on large graph datasets. Detailed descriptions of these graph datasets are provided in the Appendix. As shown in Table~\ref{tab:large_data}, our proposed method is consistently superior to other methods across all settings. Additionally, on certain datasets, labeled nodes are sparse \textit{e.g.,} WikiCS that contains only $4.96\%$ labeled nodes or Physics that contains only $1.45\%$. The results indicate that our method is robust even in the presence of a small number of labeled nodes.

\begin{table*} [!h]
    \caption{Mean and standard deviations of classification accuracy (percentage) on large graph datasets with instance-dependent label noise. The results are the mean over five trials and the best are bolded.}
    \vspace{-5pt}
	\centering
\scriptsize
	\resizebox{1\linewidth }{!}{
         \midsepremove

	\begin{tabular}{c |cc|cc|cc|cc}
		\toprule
             Dataset & \multicolumn{2}{c|}{\textit{WikiCS}} & \multicolumn{2}{c|}{\textit{Facebook}} & \multicolumn{2}{c|}{\textit{Physics}} & \multicolumn{2}{c}{\textit{DBLP}}\\
            \hline

	 Method & 30\% & 50 \%& 30\% &  50\% & 30\% & 50\% &  30\% & 50\%\\
		\hline	

		CP  & 72.27±0.40 	& 54.41±1.75 &	74.86±1.19 &	62.46±3.47	& 90.64±1.38 & 81.88±0.96 &	70.02±3.06 & 55.54±5.58\\
  
  		NRGNN & 73.09±1.63	&	56.10±2.67	& 68.00±2.34 &	58.34±3.69 &	88.96±2.23	& 82.04±1.06 & 72.48±2.61 & 65.42±9.63	\\

             PI-GNN & 75.28±0.56	& 58.51±1.24 &	 75.18±0.26 &		60.32±0.26 & 89.16±1.03 &  82.14±0.94 &	71.72±3.39 & 62.31±2.26\\

		Co-teaching+ & 72.64±0.81 & 54.66±2.18 & 75.19±1.53 & 60.48±3.22 & 90.08±1.71 & 78.07±4.73 & 66.32±2.12 & 51.46±4.49\\

		Me-Momentum  & 75.75±0.28 & 58.40±1.95 &62.86±1.39 &  46.13±1.67 & 82.65±0.69 & 68.22±2.47 & 59.88±0.60 &  44.54±2.34\\
    
     MentorNet & 72.17±0.98 &  51.80±3.30 & 73.74±2.07 & 59.04±3.38 & 88.59±2.51 & 76.31±4.50 & 63.73±4.93 & 47.85±6.47 \\

      CLNode & 73.98±0.40 &  58.93±1.12 & 77.14±2.35 & 59.08±2.63 & 90.96±1.14 & 80.89±2.36 & 72.32±2.06 & 61.21±3.07 \\

        RCL & 64.88±0.72 &  55.14±0.01 & 67.20±0.01 & 52.70±1.04 & 85.16±1.34 & 72.14±1.72 & 63.20±0.81 & 48.12±1.16 \\

\hline
         \rowcolor{LightGreen}
        TSS & \textbf{76.35±0.06} & \textbf{59.33±0.46} & \textbf{77.58±1.81} & \textbf{64.46±1.75} & \textbf{92.64±0.82} &  \textbf{86.04±1.03} & \textbf{74.70±1.72} & \textbf{66.30±1.13}\\

		\bottomrule
	\end{tabular} 
 }
	\label{tab:large_data}
	\vspace{-10pt}
\end{table*}

\vspace{-5pt}
\paragraph{Hyperparameter sensitivity}
In TSS, the hyperparameter $\lambda$ affects the performance by controlling the construction of each selection. Correspondingly, the pacing function $\lambda(t)$ with training epoch number $t$ controls the increasing speed of $\lambda$, while $\lambda_{0}$ controls the initial number of $\lambda$~\citep{wang2021survey}. Thus, we evaluate the sensitivity of TSS to $\lambda(t)$ and $\lambda_{0}$. From Fig.~\ref{figure::hyper} (a), We find that the performance is relatively similar when applying different pacing functions. Additionally, the results in Fig.~\ref{figure::hyper} (b) show the performance is relatively good when $\lambda_{0}$ is between 0.3 and 0.7. 

\begin{figure*} [t]
\centering
\includegraphics[width=1\linewidth]{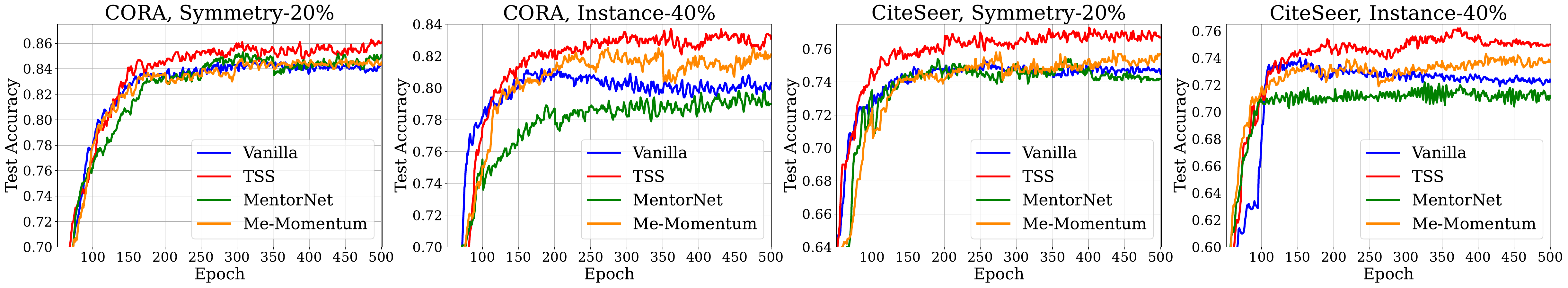}
\caption{Illustration the effectiveness of TSS on noisy \textit{CORA} and \textit{CiteSeer}. ``Vanilla" as a curriculum learning is based on the straightforward selection with confidence, instead of the CBC measure.}
\label{figure::acc_curve}

\end{figure*}

\begin{table*} [t]
	\caption{Mean and standard deviations of classification accuracy (percentage) on different GNN architectures. The experimental results are reported over five trials. Bold numbers are superior results.}
	\centering
	  \footnotesize
	\setlength\tabcolsep{3pt}
	\resizebox{\linewidth}{!}{
         \midsepremove
	\begin{tabular}{l |cc|cc|cc|cc}
		\toprule
             Dataset & \multicolumn{4}{c|}{\textit{CORA}} & \multicolumn{4}{c}{\textit{CiteSeer}} \\
            \hline
 
          \multirow{2}{*}{Backbone} & \multicolumn{2}{c|}{\textit{Symmetric}} & \multicolumn{2}{c|}{\textit{Pairflip}} & \multicolumn{2}{c|}{\textit{Symmetric}} & \multicolumn{2}{c}{\textit{Pairflip}}\\
          \cline{2-9} 
	 & 20\% & 40 \%& 20\% &  40\% & 20\% & 40\% &  20\% & 40\%\\
		\hline	
     GCN & 85.96±0.22 & 82.58±0.92 & 85.26±0.30 & 65.15±1.53 & \textbf{76.87±0.37} &74.77±0.79 & \textbf{77.25±0.44} & \textbf{65.36±1.27} \\
	\hline	
      GAT & 86.12±0.50 & \textbf{82.68±0.78} & 85.86±0.44 & 66.26±0.79 &76.16±0.40 & 73.72±0.19 & 76.98±0.30 & 64.48±1.63\\
      \hline	
       ARMA & 85.82±0.40 & 81.32±0.80 & 84.20±0.27 & 65.48±1.11& 75.22±0.37 & 72.80±0.60 & 75.32±0.78 & 63.86±1.17 \\
	\hline	
       APPNP & \textbf{86.54±0.45} & 82.20±0.68 & \textbf{86.20±0.44} & \textbf{66.64±1.00} & 76.70±0.26 & \textbf{75.66±0.44} &76.64±0.45& 65.32±1.69\\

		\bottomrule
	\end{tabular}
 }
	\label{tab:GNN_architectures}
\end{table*}

\begin{table*} [!h]
	\caption{Mean and standard deviations of classification accuracy (percentage) on different difficulty measurer. The experimental results are reported over five trials. Bold numbers are superior results.}
	\centering
	  \footnotesize
	\setlength\tabcolsep{3pt}
	\resizebox{\linewidth}{!}{
        \midsepremove
	\begin{tabular}{l |ll|ll|ll|ll}
		\toprule
             Dataset & \multicolumn{4}{c|}{\textit{CiteSeer}} & \multicolumn{4}{c}{\textit{PubMed}} \\
            \hline
 
          \multirow{2}{*}{Difficulty Measurer} & \multicolumn{2}{c|}{\textit{Symmetric}} & \multicolumn{2}{c|}{\textit{Instance-dependent}} & \multicolumn{2}{c|}{\textit{Symmetric}} & \multicolumn{2}{c}{\textit{Instance-dependent}}\\
          \cline{2-9} 
	 &  \makecell[c]{30\%} & \makecell[c]{50\%} & \makecell[c]{30\%} &  \makecell[c]{50\%} & \makecell[c]{30\%} & \makecell[c]{50\%} &  \makecell[c]{30\%} & \makecell[c]{50\%}\\
		\hline	
     Feature-based & 74.35±0.86 & 68.77±0.59 & 74.50±0.16	& 70.30±0.12 & 84.11±0.76	&	81.64±0.50 & 84.10±0.04	&	81.72±0.06 \\
	\hline	
      Neighborhood-based & 74.54±0.36 &	68.93±0.78 & 74.72±0.10	&	68.90±0.11 &84.15±0.88	&	81.86±0.43 & 84.28±0.23	&	81.76±0.10\\
      \hline	
    \rowcolor{LightGreen}
        CBC-based & \textbf{75.86±0.31} 	&	\textbf{71.81±0.74}  & \textbf{76.61±0.17} 	&	\textbf{74.03±0.26}  & \textbf{86.69±0.32} 	&	\textbf{83.53±0.23}  & \textbf{86.21±0.03} 	&	\textbf{83.94±0.08}  \\
		\bottomrule
	\end{tabular}
 }
	\label{tab:diff_measasurer}
\vspace{-10pt}
\end{table*}

\subsection{Ablation Study}

\paragraph{Performance with different GNN architectures}
We evaluate our proposed TSS on different GNN architectures, i.e., GCN~\citep{zhang2019graph}, GAT~\citep{velivckovic2017graph}, ARMA~\citep{bianchi2021graph} and APPNP~\citep{gasteiger2018predict}. The experiments are conducted on Cora and CiteSeer datasets, which are shown in Table~\ref{tab:GNN_architectures}. As can be seen, TSS performs similarly on different GNN architectures, showing consistent generalization on different architectures.

\paragraph{Performance with different difficulty measurers}
We compare our proposed CBC measurement with other two baseline measurements: The feature-based difficulty measurer and the neighborhood-based difficulty measurer in Table~\ref{tab:diff_measasurer}. The results clearly demonstrate the enhanced performance of the CBC-based difficulty measurer. Notably, the extent of accuracy improvement presents a consistent upward trend as the noise rate increases. This observation further underscores the efficacy and value of the CBC-based approach in effectively dealing with label noise.
\vspace{-5pt}
\paragraph{The underlying mechanism of TSS}
To assess whether the ``easy-to-hard" mechanism of TSS effectively extracts informative nodes, we design an \textit{vanilla} method that extracts the confident nodes once at the beginning of training epochs and trains a GNN on the totally extracted nodes during all epochs. The initial extraction process is similar to TSS. From the comparison in the Fig.~\ref{figure::acc_curve}, we can see that the TSS gradually improves the training efficiency by introducing more informative nodes and reaches better performance than the vanilla method. Additionally, the utilization of two baseline sample selection methods further demonstrates the effectiveness of our approach. This proves the necessity of introducing the ``easy-to-hard" learning schedule along with CBC to alleviate the poor extraction performance from informative nodes during the cold-start stage.

\vspace{-5pt}
\section{Conclusion}

To handle the challenge of extracting clean nodes on the noisily labeled graph, we propose a \textit{Topological Sample Selection} (TSS) method that exploits the topological information to boost the informative sample selection process. TSS utilizes the proposed \textit{Class-conditional Betweenness Centrality} (CBC) measure to characterize the topological structure of each node, steering the model to initially extract and learn from the nodes situated away from class boundaries. Subsequently, TSS focuses on extracting clean informative nodes near class boundaries. This improved sample selection process significantly enhances the robustness of the trained model against label noise. The effectiveness of this method has been proved by our theoretical analysis and extensive experiments. In the future, we will continually explore the robustness of TSS for other imperfect graph data, for example, imbalanced graph data or out-of-distribution graph data to demonstrate its effectiveness.

\section*{Acknowledgements}

TLL is partially supported by the following Australian Research Council projects: FT220100318, DP220102121, LP220100527, LP220200949, and IC190100031. JCY is supported by the National Key R\&D Program of China (No. 2022ZD0160703), 111 plan (No. BP0719010) and National Natural Science Foundation of China (No. 62306178). BH is supported by the NSFC General Program No. 62376235 and Guangdong Basic and Applied Basic Research Foundation Nos. 2022A1515011652 and 2024A1515012399. The authors would give special thanks to Muyang Li for helpful discussions and comments. The authors thank the reviewers and the meta-reviewer for their helpful and constructive comments on this work.

\section*{Impact Statement}

Noisy labels have become prevalent in the era of big data, posing significant reliability challenges for traditional supervised learning algorithms. The impact of label noise is even more pronounced in graph data, where the noise can propagate through topological edges. Effectively addressing noisy labels on graph data is a critical issue that significantly impacts the practical applications of graph data, garnering increasing attention from both the research and industry communities. 

In this study, we introduce a Topological Sample Selection (TSS) framework to mitigate the adverse effects of label noise by selectively extracting nodes with clean labels. The effectiveness of TSS is supported by substantial evidence detailed in the paper. The outcomes of this research will advance our understanding of handling label noise in graph data and substantially enhance the robustness of graph models, making strides toward more reliable and accurate graph-based learning.

\medskip
\bibliography{main}

\clearpage

\onecolumn
\appendix

\etocdepthtag.toc{mtappendix}
\etocsettagdepth{mtchapter}{none}
\etocsettagdepth{mtappendix}{subsection}

\renewcommand{\contentsname}{Appendix}
\tableofcontents
\section{A Further Discussion on Class-conditional Betweenness Centrality}
\label{CBC_diss}

\subsection{Background of Betweenness Centrality}
When shaping classifiers by GNNs in graph-structured data, some nodes situated near topological class boundaries are important to drive the decision boundaries of the trained classifier~\citep{chen2021topology}. However, GNNs find it challenging to discern class characteristics from these nodes due to their aggregation of characteristics from various classes, causing them to lack the distinctive features typical of their corresponding classes~\citep{wei2023clnode}. Moreover, this heterogeneous aggregation makes it difficult to extract clean label nodes from those near the boundaries. Thus, we design a Class-conditional Betweenness Centrality (CBC) measure that can effectively detect those nodes. 

Our Class-conditional Betweenness Centrality measure is inspired by the classical concept in graph theory – Betweenness Centrality (BC). The formal definition of the Betweenness Centrality is as follows.

\begin{definition}[Betweenness centrality]
The betweenness centrality (BC) of the node $\mathbf{v}_i$ is defined to be the fraction of shortest paths between pairs of vertices in a graph $\mathcal{G}$ that pass through $\mathbf{v}_i$. Formally, the betweenness centrality of a node $\mathbf{v}_i$ is defined:
\begin{equation} \label{bc}
\mathbf{b}_{\mathbf{v}_i} = \frac{1}{n(n-1)}\sum_{\mathbf{v}_u \neq\mathbf{v}_i \neq\mathbf{v}_v} \frac{\sigma_{\mathbf{v}_u,\mathbf{v}_v}(\mathbf{v}_i)}{\sigma_{\mathbf{v}_u,\mathbf{v}_v}}
\end{equation}
where $\sigma_{\mathbf{v}_u,\mathbf{v}_v}$ denotes the number of shortest paths from $\mathbf{v}_u$ to $\mathbf{v}_v$, and $\sigma_{\mathbf{v}_u,\mathbf{v}_v}(\mathbf{v}_i)$ denotes the
number of shortest paths from $\mathbf{v}_u$ to $\mathbf{v}_v$ that pass through $\mathbf{v}_i$.
\end{definition}

\subsection{Difference of Class-conditional Betweenness Centrality}

\begin{figure*} [h]
\centering
  \includegraphics[width=8cm,height=3cm]{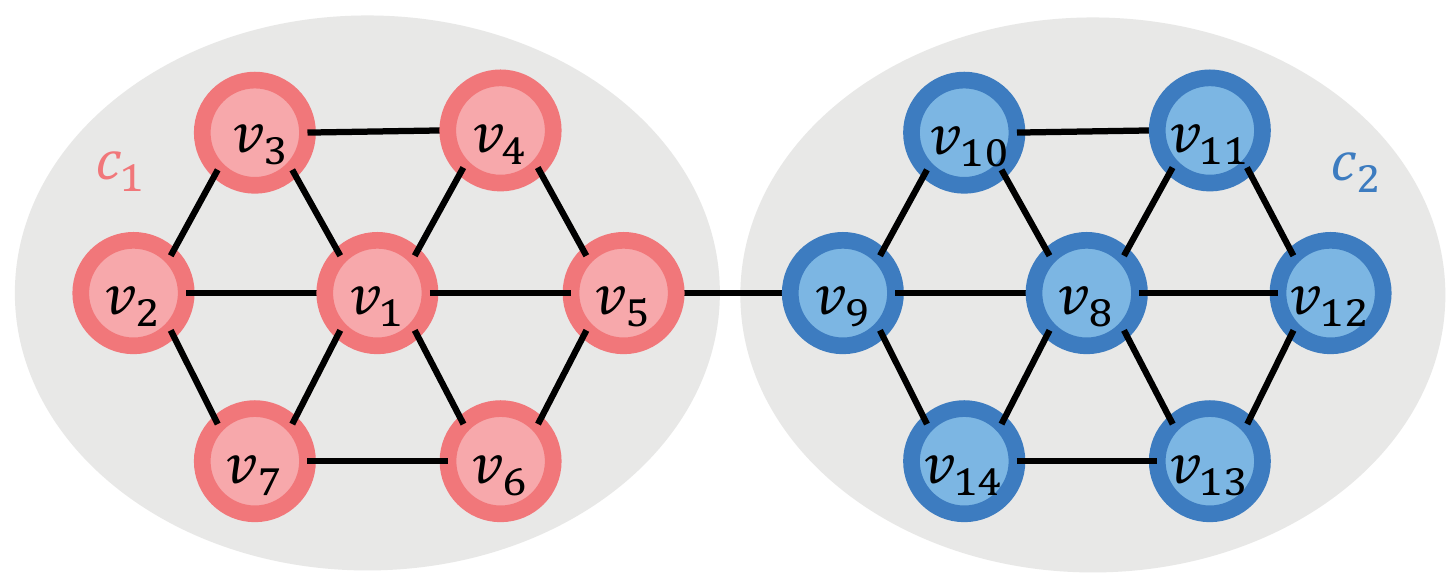}

\caption{The illustration of boundary-near nodes}
\label{figure::CBC_diss}
\end{figure*}

The betweenness centrality measures the centrality of nodes in a connected graph based on the shortest paths of other pairs of nodes. It provides a quantified measure of a node's influence in controlling the flow of information among other nodes. A higher betweenness centrality signifies a node's increased significance in regulating the information flow within the network. By incorporating the class-conditional constraint into Eq.~(\ref{bc}), we can effectively identify nodes that play a crucial role in controlling the flow of information between different classes and are typically located near topological class boundaries. This is exemplified by the boundary-near nodes $v_5$ and $v_9$ in Fig.~\ref{figure::CBC_diss}, where the shortest paths for nodes in class 1 and class 2 must pass through these nodes, underlining their pivotal role in managing information flow between the two classes.

Thus, after adding the class-conditional constrain into the Eq.(\ref{bc}), we define the CBC of a node $\mathbf{v}_i$ as the fraction of shortest paths between pairs of nodes that belong to different classes in a graph $\mathcal{G}$ that pass through $\mathbf{v}_i$:
\begin{equation} \label{formal_CBC}
\mathbf{Cb}_{i} = \frac{1}{n(n-1)}\sum_{\substack{\mathbf{v}_u \neq\mathbf{v}_i \neq\mathbf{v}_v \\ y_{u} \neq y_{v}}}
\frac{\sigma_{\mathbf{v}_u,\mathbf{v}_v}(\mathbf{v}_i)}{\sigma_{\mathbf{v}_u,\mathbf{v}_v}}
\end{equation}
where $\sigma_{\mathbf{v}_u,\mathbf{v}_v}$ denotes the number of shortest paths from $\mathbf{v}_u$ to $\mathbf{v}_v$, and $\sigma_{\mathbf{v}_u,\mathbf{v}_v}(\mathbf{v}_i)$ denotes the
number of shortest paths from $\mathbf{v}_u$ to $\mathbf{v}_v$ that pass through $\mathbf{v}_i$. 

Notably, the CBC measure builds upon the BC measure and outperforms it in detecting boundary-near nodes. This improvement is attributed to the class-conditional constraint, which alleviates the impact of information flow among nodes belonging to the same class. Specifically, information flow among nodes of the same class is more likely to occur through nodes positioned near the centre of the class rather than at the boundary. For instance, in Fig.~\ref{figure::CBC_diss}, the shortest path from node $v_3$ to $v_6$ or from $v_7$ to $v_4$ traverses the centre-near node $v_1$ rather than the boundary-near node $v_5$. 

\subsection{Optimization Form of Class-conditional Betweenness Centrality}

However, it is usually practically limited to directly employ Eq.~(\ref{formal_CBC}), since in most networks, the information does not flow only along the shortest paths~\citep{stephenson1989rethinking,freeman1991centrality,newman2005measure}, and it is very time-consuming to find the shortest paths in a large graph~\citep{liu2010link,zhao2022random}. 
Thus, we relax Eq.~(\ref{formal_CBC}) with the \textit{random walk}, which simultaneously allows the multiple paths to contribute to CBC and avoids the expensive search cost of the shortest paths~\citep{noh2004random,liu2010link,zhao2022random}. Concretely, we employ the Personalized PageRank (PPR) method~\citep{bahmani2010fast,haveliwala2003analytical} to implement random walk and then arrive at the final form of our CBC in the following definition.

\begin{definition}[Class-conditional Betweenness Centrality] 
Given the Personalized PageRank matrix $\boldsymbol{\pi} = \alpha(\mathbf{I}-(1-\alpha)\hat{\mathbf{A}})^{-1}$ ($\boldsymbol{\pi} \in \mathbb{R}^{n \times n}$), the Class-conditional Betweenness Centrality of the node $\mathbf{v}_i$ is defined by counting how often the node $\mathbf{v}_i$ is traversed by a random walk between pairs of other vertices that belong to different classes in a graph $\mathcal{G}$:
\begin{equation} \label{RCBC_equation}
\mathbf{Cb}_{i} \vcentcolon=  \frac{1}{n(n-1)}\sum_{\substack{\mathbf{v}_u \neq\mathbf{v}_i \neq\mathbf{v}_v \\ \tilde{y}_u \neq \tilde{y}_v}}
\frac{\boldsymbol{\pi}_{u,i}\boldsymbol{\pi}_{i,v}}{\boldsymbol{\pi}_{u,v}},
\end{equation}
where $\boldsymbol{\pi}_{u,i}$ with the target node $\mathbf{v}_u$ and the source node $\mathbf{v}_i$ denotes the probability that an $\alpha$-discounted random walk from node $\mathbf{v}_u$ terminates at $\mathbf{v}_i$. Here an $\alpha$-discounted random walk represents a random traversal that, at each step, either terminates at the current node with probability $\alpha$, or moves to a random out-neighbour with probability $1 - \alpha$.
\end{definition}

In the above definition, the CBC is based on the random walks that count how often a node is traversed by a random walk between pairs of other nodes that belong to different classes. Our proposed CBC successfully detects the boundary-near nodes by evaluating the flow of messages passing between different classes. The nodes that possess high CBC are closer to the topological class boundaries. Consequently, our CBC measure is adept at identifying the topological structure of nodes, and its exploration of topological information renders it robust against noisy labeled data. Additionally, the CBC measure can be seamlessly integrated into other related domains. For instance, it can be employed to identify the structure of nodes in out-of-distribution (OOD) detection tasks, as discussed in~\cite{wu2022energy,huang2023harnessing}, and to enhance OOD generalization, as demonstrated in studies by~\cite{yang2022geometric,huang2023robust,peng2023sam} and~\cite{wu2021handling,huang2023winning}.

\subsection{Importance of Class-conditional Betweenness Centrality}

\begin{figure*} [h]
\centering
  \includegraphics[width=0.4\linewidth,height=5.0cm]{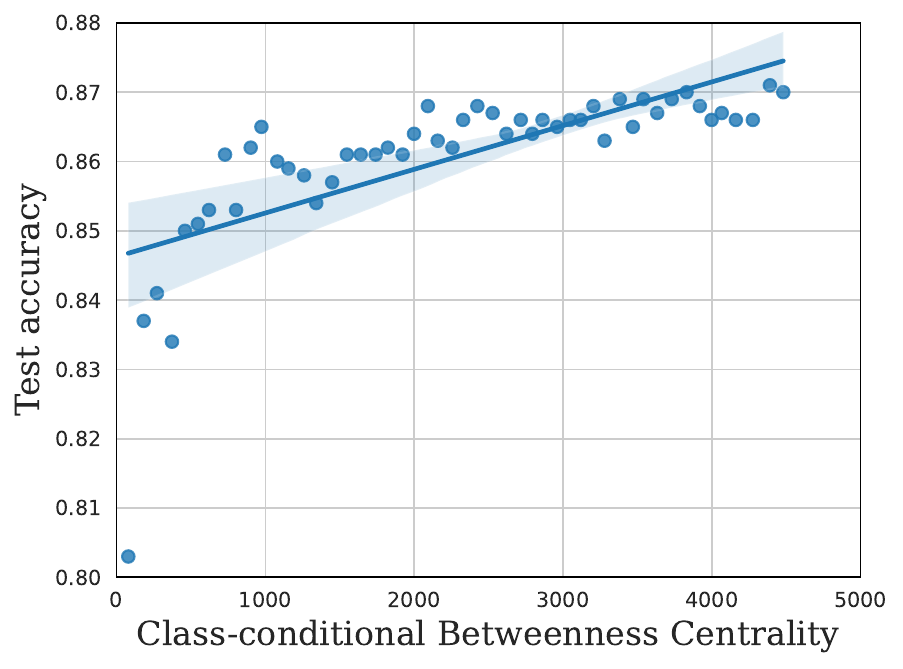}

\caption{There is a significant positive correlation between the test accuracy and the overall CBC of the clean labeled training set (the Pearson correlation coefficient is $0.6999$ over 50 randomly selected class-balanced training sets with the $p$ value smaller than 0.0001).}
\label{figure::CBC}
\end{figure*}

In Fig.~\ref{figure::CBC}, we present a visual representation highlighting the clear positive correlation between test accuracy and the aggregate Class-conditional Betweenness Centrality (CBC) of the training set. Additionally, we carefully structure the training sequence for each node in every training set, prioritizing nodes based on their CBC scores. This underlines the pivotal role of CBC in shaping the performance of models. The empirical findings strongly affirm the significance of extracting insights from informative nodes, a factor that markedly enhances the performace of GNNs throughout the training process.

\subsection{Distribution of Class-conditional Betweenness Centrality}

\begin{figure*} [h] 
\centering
\includegraphics[width=1\linewidth]{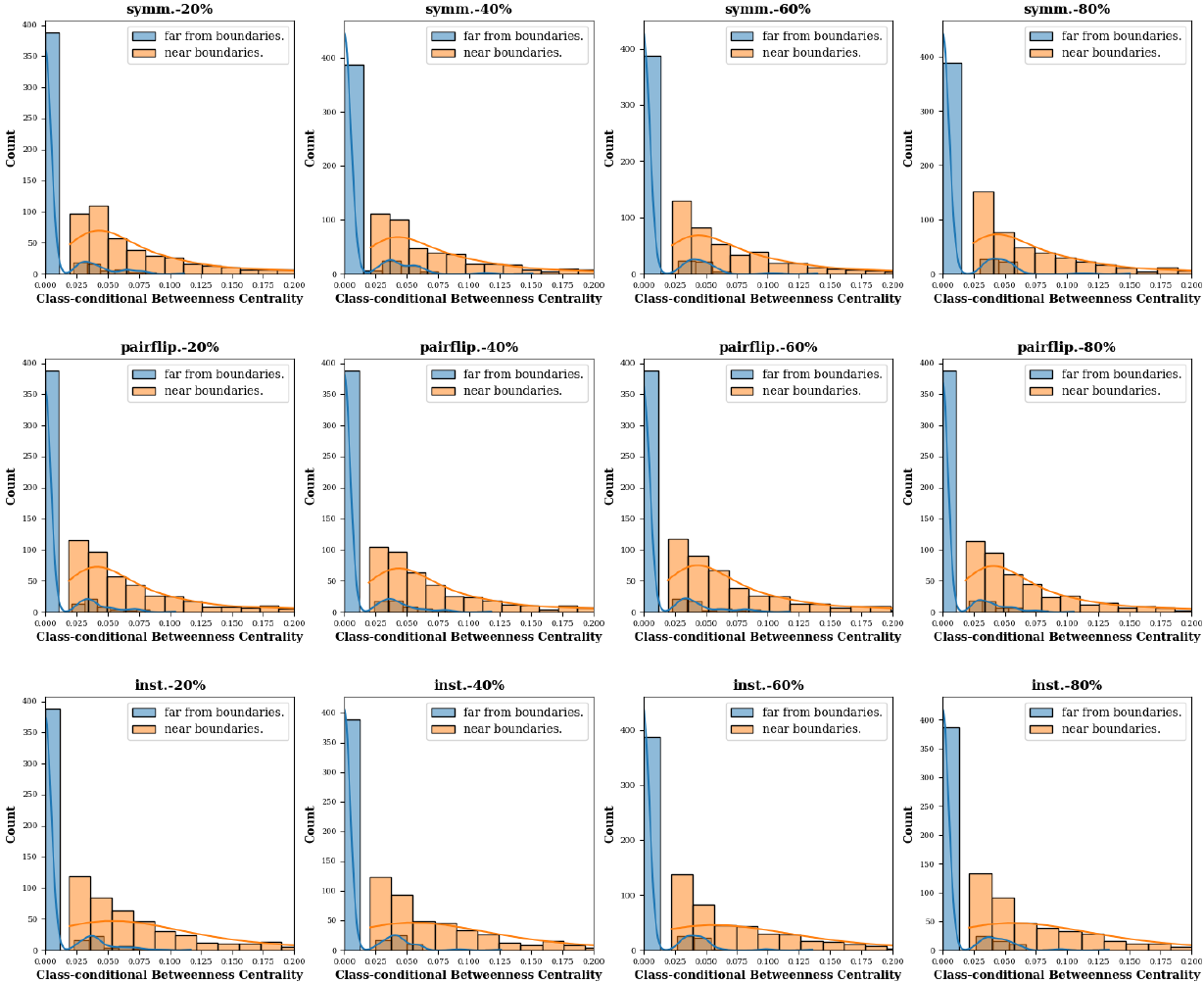}
\caption{Class-conditional Betweenness Centrality distributions of nodes in WikiCS, with varying levels of symmetric noise (symm.), pairflip noise (pairflip.), and instance-based noise (inst.).}
\label{figure::CBC_dis_total}
\end{figure*}

In our comprehensive empirical analysis, we thoroughly investigate the distributions of Class-conditional Betweenness Centrality for nodes in WikiCS, considering diverse levels of noise as presented in Fig.~\ref{figure::CBC_dis_total}. To pre-categorize nodes based on their proximity to topological class boundaries, we employ the following criteria: Nodes are classified as ``far from topological class boundaries" (far from boundaries) if their two-hop neighbors belong to the same class. Conversely, nodes are labeled as ``near topological class boundaries" (near boundaries) if this condition does not apply. It's important to note that the ``WikiCS" dataset, chosen for this analysis, is substantial and comprises sparsely labeled nodes. As observed in Fig.~\ref{figure::CBC_dis_total}, the node dataset exhibits two distinct clusters. Even in the presence of considerable label noise, nodes far away from topological class boundaries consistently demonstrate lower CBC scores across all cases.

\section{Related work} \label{related_work}

\subsection{Curriculum Learning with Label Noise}

We have diligently incorporated curriculum-based approaches into our literature review that align with our research theme. One widely adopted criterion involves selecting samples with small losses and treating them as clean data. Several curriculum learning methods utilize this criterion~\citep{jiang2014self}, and in each step, select samples with small losses. For instance, in MentorNet~\citep{jiang2018mentornet}, an additional pre-trained network is employed to select clean instances using loss values to guide model training. The underlying concept of MentorNet resembles the self-training approach~\citep{kumar2010self}, inheriting the drawback of accumulated error due to sample-selection bias. 

To address this issue, Co-teaching~\citep{han2018co} and Co-teaching+~\citep{yu2019does} mitigate the problem by training two DNNs and using the loss computed on one model to guide the other. CurriculumNet~\citep{guo2018curriculumnet}presents a curriculum learning approach based on unsupervised estimation of data complexity through its distribution in a feature space. It benefits from training with both clean and noisy samples and weights each sample's loss in training based on the gradient directions compared to those on validation (\ie, a clean set). Notably, CurriculumNet relies on a clean validation set. 

It's worth emphasizing that the discussed curriculum learning methods primarily focus on mitigating label noise issues within i.i.d. datasets and depend on the prediction of pre-trained neural networks. However, those methods cannot be employed on graph data due to the ``over-smoothing" issue when training Graph Neural Networks (GNNs). Note that, in GNNs, ``over-smoothing" refers to the phenomenon where, as the network depth increases, node features become increasingly similar. This similarity poses a challenge when employing curriculum learning with label noise, making it difficult to distinguish between ``easy" and ``hard" nodes due to the homogenization of features caused by over-smoothing. Additionally, even in shallow GNN architectures, over-smoothing can lead to under-confident predictions, complicating the task of establishing an 'easy-to-hard' training curriculum ~\citep{wang2021confident,hsu2022makes}. Addressing this challenge, our work introduces a novel method, which proposes a robust CBC measure. This measure effectively distinguishes between 'easy' and 'hard' nodes, taking into account the graph structure rather than the prediction of GNNs, thereby mitigating the over-smoothing problem. Our work stands as a pioneer in the development of a curriculum learning approach explicitly designed for graph data afflicted by label noise. This distinction underscores a significant contribution of our research, emphasizing the necessity for specialized strategies to effectively handle noise within graph-structured data.

\subsection{Graph Neural Netwoeks}

Predicting node labels involves formulating a parameterized hypothesis using the function 	$f_{\mathcal{G}}(\mathbf{A}, \mathcal{X}) = {\hat{y}}_{\mathbf{A}}$, incorporating a Graph Neural Network (GNN) architecture~\citep{kipf2016semi} and a message propagation framework~\citep{gilmer2017neural}. The GNN architecture can take on various forms such as GCN~\citep{kipf2016semi}, GAT~\citep{velivckovic2017graph}, or GraphSAGE~\citep{hamilton2017inductive}.

In practical terms, the forward inference of an $L$-layer GNN involves generating node representations 	$\bm{H}_{\mathbf{A}} \in \mathbb{R}^{N \times D}$  through $L$-layer message propagation. Specifically, with $\ell=1\dots L$ denoting the layer index, $h_{i}^{\ell}$ is the representation of the node $i$, $\text{MESS}(\cdot)$ being a learnable mapping function to transform the input feature, $\text{AGGREGATE}(\cdot)$ capturing 1-hop information from the neighborhood $\mathcal N(v)$ in the graph, and $\text{COMBINE}(\cdot)$ signifying the final combination of neighbor features and the node itself, the $L$-layer operation of GNNs can be formulated as $
	\bm m_v^{\ell} \! = \! \text{AGGREGATE}^{\ell}(\{ \text{MESS}(\bm h_u^{\ell-1}, \bm h_v^{\ell-1}, e_{uv}) \! :\!  u \in \mathcal{N}(v)\})
	$, where $\bm h_v^{\ell} = \text{COMBINE}^{\ell}(\bm h_v^{\ell-1}, \bm m_v^{\ell})$. After L-layer propagation, the final node representations $\bm h_{e}^L$ for each $e  \in  V$ are derived. Furthermore, a detailed summary of different GNN architectures is presented in Table~\ref{tab: GNN-architectures}.

Subsequently, a subsequent linear layer transforms $\bm{H}_{\mathbf{A}}$ into classification probabilities ${\hat{y}}_{\mathbf{A}}  \in  \mathbb{R}^{N \times C}$, where $C$ represents the total categories. The primary training objective is to minimize the classification loss, typically measured by cross-entropy between the predicted ${\hat{y}}_{\mathbf{A}}$ and the ground truth $Y$.

	\begin{table}[H]
		\centering
		\caption{Detailed architectures of different GNNs.}
		\fontsize{10}{16}\selectfont
		\setlength\tabcolsep{6pt}
		\begin{tabular}{c|c|c}
			\toprule
			GNN & $\text{MESS}(\cdot) \; \& \; \text{AGGREGATE}(\cdot)$ & $\text{COMBINE}(\cdot)$ \\ \hline
			GCN & $\bm m_{i}^{l} = \bm W^{l} \sum_{j \in \mathcal{N}(i)} \frac{1}{\sqrt{\hat{d}_i \hat{d}_j }} \bm h_{j}^{l-1}$ 
			& $\bm h_{i}^{l}  = \sigma( \bm m_{i}^{l} + \bm W^{l} \frac{1}{\hat{d}_i} \bm h_{i}^{l-1} )$  \\
			GAT & $\bm m_{i}^{l} =  \sum_{j \in \mathcal{N}(i)} \alpha_{ij} \bm W^{l} \bm h_{j}^{l-1}$ 
			& $\bm h_{i}^{l}  = \sigma( \bm m_{i}^{l} + \bm W^{l}  \alpha_{ii} \bm h_{i}^{l-1} )$  \\
			GraphSAGE & $\bm m_{i}^{l} =  \bm W^{l}  \frac{1}{|\mathcal{N}(i)|} \sum_{j \in \mathcal{N}(i)} \bm h_{j}^{l-1}$ 
			& $\bm h_{i}^{l}  = \sigma( \bm m_{i}^{l} + \bm W^{l}  \bm h_{i}^{l-1} )$  \\
			\bottomrule
		\end{tabular}
		\label{tab: GNN-architectures}
	\end{table}

\subsection{Denoising Methods on Graph Data}

Prior research has explored diverse strategies to address the challenge of label noise in graph data. NRGNN~\citep{dai2021nrgnn} combats label noise by linking unlabeled nodes with noisily labeled nodes that share high feature similarity, thus incorporating more reliable label information. Conversely, PI-GNN~\citep{du2021pi} mitigates noise impact by introducing Pairwise Intersection (PI) labels based on feature similarity among nodes.

In a different approach, the LPM method~\citep{xia2020towards} and GNN-Cleaner~\citep{xia2023gnn} address noisy labels by involving a small set of clean nodes for assistance. Additionally, CP~\citep{zhang2020adversarial} operates with class labels derived from clustering node embeddings, encouraging the classifier to capture class-cluster information and avoid overfitting to noisy labels.

Furthermore, RS-GNN~\citep{dai2022towards} focuses on enhancing GNNs' robustness to noisy edges. It achieves this by training a link predictor on noisy graphs, aiming to enable effective learning from graphs that contain inaccuracies in edge connections.

Lastly, RT-GNN~\citep{qian2023robust} leverages the memorization effect of neural networks to select clean labeled nodes, generating pseudo-labels from these selected nodes to mitigate the influence of noisy nodes on the training process. 

In addition, the efficacy of contrastive learning~\citep{you2020graph,chen2023clip2scene,zheng2022towards,zheng2023enhancing} has been harnessed to effectively reduce label noise during node classification tasks on graph-based data. Based on the homophily assumption, ALEX~\citep{yuan2023alex} learns robust node representations utilizing graph contrastive learning to mitigate the overfitting of noisy nodes and CGNN~\citep{yuan2023learning} integrates graph contrastive learning as a regularization term, thereby bolstering the robustness of trained models against label noise. Each of these approaches offers unique insights into effectively handling label noise in graph data. 

In this context, our proposed Topological Sample Selection(TSS) represents a distinctive perspective on employing curriculum learning methods specifically tailored for noisily labeled graphs. By introducing TSS, we contribute a novel and effective strategy to tackle label noise in the complex domain of graph-structured data.

\subsection{Graph Curriculum Learning}

Graph Curriculum Learning (GCL) stands at the intersection of graph machine learning and curriculum learning, gaining increasing prominence due to its potential. At its core, GCL revolves around customizing a difficulty measure to compute a difficulty score for each data sample, crucial in defining an effective learning curriculum for the model. The design of this difficulty measure can follow predefined or automatic approaches.

Predefined approaches often employ heuristic metrics to measure node difficulty based on specific characteristics even before the training commences. For example, CLNode~\citep{wei2023clnode} gauges node difficulty by considering label diversity among a node's neighbors. Conversely, SMMCL~\citep{gong2019multi} assumes varying difficulty levels among different samples for propagation, advocating an easy-to-hard sequence in the curriculum for label propagation.

On the other hand, automatic approaches determine difficulty during training using a supervised learning paradigm rather than predefined heuristic-based metrics. For example, RCL~\citep{zhang2023relational} gradually incorporates the relation between nodes into training based on the relation's difficulty, measured using a supervised learning approach. Another instance, MentorGNN~\citep{zhou2022mentorgnn}, tailors complex graph signals by deriving a curriculum for pre-training GNNs to learn informative node representations and enhance generalization performances.

However, a notable gap exists in existing GCL methods concerning their robustness to label noise, especially in effectively handling graphs with noisy labels. Our proposed Topological Sample Selection(TSS) addresses this limitation by being the pioneer in curriculum learning explicitly designed for graphs affected by label noise. This underscores the novelty and significance of TSS within the domain of GCL.

\section{Proof to Theoretical Guarantee of TSS}
\label{theoretical_proof}

\subsection{Proof for the Weighted Expression}

We first formulate $\mathbbm{P}_{{\mathcal{D}}}(\mathbf{A}, \mathbf{x})$ as the weighted expression of $\mathbbm{P}_{{\hat{\mathcal{D}}}}(\mathbf{A}, \mathbf{x})$:
\begin{equation} \label{simulated_curriculum_format}
    \mathbbm{P}_{{\mathcal{D}}}(\mathbf{A}, \mathbf{x}) = \frac{1}{\alpha^{*}}W_{\lambda^{*}}(\mathbf{A}, \mathbf{x})\mathbbm{P}_{{\hat{\mathcal{D}}}}(\mathbf{A}, \mathbf{x}),
\end{equation}
where $0 \leq W_{\lambda^{*}}(\mathbf{A}, x) \leq 1$ and $\alpha^{*} = \int_{\mathbf{A},\mathcal{X}}W_{\lambda^{*}}(\mathbf{A}, \mathbf{x})\mathbbm{P}_{{\hat{\mathcal{D}}}}(\mathbf{A}, \mathbf{x})dx$ denote the normalization factor. Based on Eq.(\ref{simulated_curriculum_format}), $\mathbbm{P}_{{\mathcal{D}}}(\mathbf{A}, \mathbf{x})$ actually corresponding to a curriculum as definied in Eq.(\ref{TSS}) under the weight function $W_{\lambda^{*}}(\mathbf{A}, \mathbf{x})$.

Eq.(\ref{simulated_curriculum_format}) can be equivalently reformulated as
\begin{equation} \label{weighted_expression}
\mathbbm{P}_{{\hat{\mathcal{D}}}}(\mathbf{A}, \mathbf{x}) = \alpha^{*}\mathbbm{P}_{{\mathcal{D}}}(\mathbf{A}, \mathbf{x}) + (1-\alpha^{*})E(\mathbf{A}, \mathbf{x}) \nonumber,
\end{equation}
where
\begin{equation}
E(\mathbf{A}, \mathbf{x}) = \frac{1}{1-\alpha^{*}}(1-W_{\lambda^{*}}(\mathbf{A}, \mathbf{x}))\mathbbm{P}_{{\hat{\mathcal{D}}}}(\mathbf{A}, \mathbf{x}) \nonumber.
\end{equation}

Here, the term $E(\mathbf{A}, \mathbf{x})$ measures the deviation from $\mathbbm{P}_{{\hat{\mathcal{D}}}}(\mathbf{A}, \mathbf{x})$ to $\mathbbm{P}_{{\mathcal{D}}}(\mathbf{A}, \mathbf{x})$. Recalling the previous empirical analysis of Fig.~\ref{cbc_difficult_selection}, extracting confident nodes from the early $\tilde{Q}_{\lambda}$ that emphasises the easy nodes works well. We define this period (corresponding to relatively small $\lambda$) as the high-confidence regions. In these high-confidence areas, $\mathbbm{P}_{{\mathcal{D}}}(\mathbf{A}, \mathbf{x})$ is accordant to the $\mathbbm{P}_{{\hat{\mathcal{D}}}}(\mathbf{A}, \mathbf{x})$ and thus $E(\mathbf{A}, \mathbf{x})$ corresponds to the nearly zero-weighted $\mathbbm{P}_{{\hat{\mathcal{D}}}}(\mathbf{A}, \mathbf{x})$ tending to be small. On the contrary, in later training criteria, the poor performance of extracting confident nodes causes that the $\mathbbm{P}_{{\hat{\mathcal{D}}}}(\mathbf{A}, \mathbf{x})$ cannot approximate the $\mathbbm{P}_{{\hat{\mathcal{D}}}}(\mathbf{A}, \mathbf{x})$ well in those low-confident regions. $E(\mathbf{A}, \mathbf{x})$ then imposes large weights on $\mathbbm{P}_{{\hat{\mathcal{D}}}}(\mathbf{A}, \mathbf{x})$, yielding the large deviation values. Combining with Definition~\ref{Topological Curriculum Learning}, we construct the below curriculum sequence for theoretical evaluation:
\begin{equation} \label{theoretical cl}
\hat{Q}_{\lambda}(\mathbf{A},\mathbf{x}) \propto W_{\lambda}(\mathbf{A},\mathbf{x}) \mathbbm{P}_{\hat{\mathcal{D}}}(\mathbf{A},\mathbf{x}), 
\end{equation}
where 
\begin{equation}
W_{\lambda}(\mathbf{A},\mathbf{x}) \propto \frac{\alpha_{\lambda}\mathbbm{P}_{{\mathcal{D}}}(\mathbf{A}, \mathbf{x}) + (1-\alpha_{\lambda})E(\mathbf{A},\mathbf{x})}
{\alpha^{*}\mathbbm{P}_{{\mathcal{D}}}(\mathbf{A}, \mathbf{x})+(1-\alpha)^{*}E(\mathbf{A},\mathbf{x})} \nonumber
\end{equation}
with $0 \leq W_{\lambda}(\mathbf{A}, \mathbf{x}) \leq 1$ through normalizing its maximal value as 1 and $\alpha_{\lambda}$ varies from 1 to $\alpha^{*}$ with increasing pace parameter $\lambda$.

\subsection{Proof of Theorem 1}

Now, we estimate the expected risk by the following surrogate~\citep{donini2018empirical}:
\begin{equation}
\mathcal{R}_\text{emp}(f_{\mathcal{G}}):= \frac{1}{n} \sum_{i=1}^{n_{\mathrm{cf}}} \mathcal{L}_{f_{\mathcal{G}}} (z_i).
\end{equation}

Let $\mathcal{F}$ be a function family mapping from $Z_{\mathbf{x}_i}$ to $[a,b]$, $\mathbbm{P}(Z_{\mathbf{x}_i})$ a distribution on $Z_{\mathbf{x}_i}$ and $S = (Z_{\mathbf{x}_1},\dots,Z_{\mathbf{x}_m})$ a set of i.i.d. samples drawn from $\mathbbm{P}$. The empirical Rademacher complexity of $\mathcal{F}$ with respect to $S$ is defined by 
\begin{equation}
    \hat{\mathfrak{R}}_{m}(\mathcal{F}) = \mathbb{E}_{\sigma}[\sup_{g \in \mathcal{F}}\frac{1}{m}\sum_{i=1}^{m}\sigma_{i}g(Z_{\mathbf{x}_i})],
\end{equation}
where $\sigma_{i}$ are i.i.d. samples drawn from the uniform distribution in $\{-1,1\}$. The Rademacher complexity of $\mathcal{F}$ is defined by the expectation of $\hat{\mathfrak{R}}_{m}$ over all samples $S$:
\begin{equation}
    {\mathfrak{R}}_{m}(\mathcal{F}) = \mathbbm{E}_{S\sim \mathbbm{P}^{m}}|\hat{\mathfrak{R}}_{S}(\mathcal{F})|.
\end{equation}

\begin{definition}
    The Kullback-Leibler divergence $D_{KL}(p || q)$ between two densities $p(\Omega)$ and $q(\Omega)$ is defined by 
    \begin{equation}
        D_{KL}(p || q) = \int_{\Omega}p(\mathbf{x})\log\frac{p(\mathbf{x})}{q(\mathbf{x})}d\mathbf{x}.
    \end{equation}
\end{definition}

Based on the above definitions, we can estimate the generalization error bound for curriculum learning under the curriculum $\hat{Q}_{\lambda}$. Based on the Bretagnolle-Huber inequality~\citep{schluter2013novel}, we have
\begin{equation}
    \int|p(\mathbf{x}) - q(\mathbf{x})|d\mathbf{x} \leq 2 \sqrt{1-\exp\{-D_{KL}(p || q)\}}
\end{equation}

Let $\mathcal{H}$ be a family of functions taking value in $\{-1,1\}$ , for any $\delta > 0$ with confidence at least $1-\delta$ over a sample set $S$, the following holds for any $f_{\mathcal{G}} \in \mathcal{H}$~\citep{gong2016curriculum}:
\begin{equation} \label{bound_BH}
    \mathcal{R}(f_{\mathcal{G}}) \leq \mathcal{R}_{emp}(f_{\mathcal{G}}) + {\mathfrak{R}}_{m}(\mathcal{H}) + \sqrt{\frac{\ln(\frac{1}{\delta})}{2m}}.
\end{equation}
In addition, we have 
\begin{equation}
    \mathcal{R}(f_{\mathcal{G}}) \leq \mathcal{R}_{emp}(f_{\mathcal{G}}) + {\mathfrak{R}}_{m}(\mathcal{H}) + 3\sqrt{\frac{\ln(\frac{1}{\delta})}{2m}}.
\end{equation}
Suppose $S \subseteq \{ \mathbf{x}: \Vert \mathbf{x} \Vert \leq R\}$ be a sample set of size $m$, and $\mathcal{H} = \{x \rightarrow sgn(\mathbf{w}^{T}\mathbf{x}): min_{s}|\mathbf{w}^{T}\mathbf{x}| = 1 \cap ||\mathbf{w}|| \leq B \}$ be hypothesis class, where $\mathbf{w} \in \mathbb{R}^{n}, \mathbf{x} \in \mathbb{R}^{n}$, and then we have 
\begin{equation}
    \hat{\mathfrak{R}}_{m}(\mathcal{H}) \leq \frac{BR}{\sqrt{m}}
\end{equation}

\begin{proof}
    \begin{equation}
    \begin{split}
    \hat{\mathfrak{R}}_{m}(\mathcal{H}) &= \frac{1}{m} \mathbb{E}_{\sigma}\bigg[\sup_{\|\mathbf{w}\|\leq B}\sum_{i=1}^{m}\sigma_{i}sgn(\mathbf{w} {\mathbf{x}}_i)\bigg]\\
    &\leq \frac{1}{m} \mathbb{E}_{\sigma}\bigg[\sup_{\|\mathbf{w}\|\leq B}\sum_{i=1}^{m}\sigma_{i}|sgn(\mathbf{w} {\mathbf{x}}_i)|\bigg]
    \leq \frac{1}{m} \mathbb{E}_{\sigma}\bigg[\sup_{\|\mathbf{w}\|\leq B}\sum_{i=1}^{m}\sigma_{i}|\mathbf{w}{\mathbf{x}}_i|\bigg]\\
    &\leq \frac{B}{m}\mathbb{E}_{\sigma}\bigg[\sum_{i=1}^{m}\sigma_{i}\|{\mathbf{x}}_i\|\bigg] 
    \leq \frac{B}{m}\mathbb{E}_{\sigma}\bigg[| \sum_{i=1}^{m}\sigma_{i}\|{\mathbf{x}}_i\| | \bigg]\\
    &= \frac{B}{m}\mathbb{E}_{\sigma}\bigg[\sqrt{( \sum_{i=1}^{m}\sigma_{i}\|{\mathbf{x}}_i\|) ^2} \bigg]\\
    & = \frac{B}{m}\mathbb{E}_{\sigma}\bigg[\sqrt{ \sum_{i,j=1}^{m}\sigma_{i}\sigma_{j}\|{\mathbf{x}}_i \| \|{\mathbf{x}}_j\|} \bigg]\\
    &\leq \frac{B}{m} \sqrt{\mathbb{E}_{\sigma}\bigg[\sum_{i,j=1}^{m}\sigma_{i}\sigma_{j}\|{\mathbf{x}}_i\|\|{\mathbf{x}}_j\|\bigg]}\\
    &= \frac{B}{m}\sqrt{\sum_{i=1}^{m}\|{\mathbf{x}}_i\|^{2}}\\
     & \leq \frac{BR}{\sqrt{m}}.\\
    \end{split}
\end{equation}
\end{proof}

Then, suppose $\{(Z_{\mathbf{x}_i},y_i)\}^{m}_{i=1}$ are i.i.d. samples drawn from the confident pace distribution $\hat{Q}_{\lambda}$. Denote $m_{+}/m_{-}$ be the number of positive/negative samples and $m^{*} = \min\{m_{-},m_{+}\}$. $\mathcal{H}$ is the function family projecting to $\{-1,1\}$. Then for any $\delta > 0$ and $f \in \mathcal{H}$, with confidence at least $1-2\delta$ we have:
\begin{equation} \label{bound_01}
    \begin{split}
    \mathcal{R}(f_{\mathcal{G}}) &\leq \frac{1}{2}\mathcal{R}^{+}_{emp}(f_{\mathcal{G}})+\frac{1}{2}\mathcal{R}^{-}_{emp}(f_{\mathcal{G}})\\
    &+\frac{1}{2}\hat{\mathfrak{R}}_{m_+}(\mathcal{H}) + \frac{1}{2}\hat{\mathfrak{R}}_{m_-}(\mathcal{H}) + \sqrt{\frac{\ln(\frac{2}{\delta})}{m^*}}\\
    &+(1-\alpha_{\lambda})\sqrt{1-\exp{\{-D_{KL}(\mathbbm{P}^{+}_{\mathcal{D}}||E^{+})\}}}\\
    &+(1-\alpha_{\lambda})\sqrt{1-\exp{\{-D_{KL}(\mathbbm{P}^{-}_{\mathcal{D}}||E^{-})\}}},
    \end{split}
\end{equation}
and 
\begin{equation} \label{bound_01}
    \begin{split}
    \mathcal{R}(f_{\mathcal{G}}) &\leq \frac{1}{2}\mathcal{R}^{+}_{emp}(f_{\mathcal{G}})+\frac{1}{2}\mathcal{R}^{-}_{emp}(f_{\mathcal{G}})\\
    &+\frac{1}{2}\hat{\mathfrak{R}}_{m_+}(\mathcal{H}) + \frac{1}{2}\hat{\mathfrak{R}}_{m_-}(\mathcal{H}) + 3\sqrt{\frac{\ln(\frac{2}{\delta})}{m^*}}\\
    &+(1-\alpha_{\lambda})\sqrt{1-\exp{\{-D_{KL}(\mathbbm{P}^{+}_{\mathcal{D}}||E^{+})\}}}\\
    &+(1-\alpha_{\lambda})\sqrt{1-\exp{\{-D_{KL}(\mathbbm{P}^{-}_{\mathcal{D}}||E^{-})\}}},
    \end{split}
\end{equation}
where $E^{+}, E^{-}$ denotes the error distribution corresponding to $\mathbbm{P}_{\mathcal{D}}(\mathbf{A},x|y = 1), \mathbbm{P}_{\mathcal{D}}(\mathbf{A},x|y = -1)$, and $\mathcal{R}^{+}_{emp}(f_{\mathcal{G}}), \mathcal{R}^{-}_{emp}(f_{\mathcal{G}})$ denote the empirical risk on positive nodes and negative nodes, respectively. 

\begin{proof}
    We first rewrite the expected risk as:
\begin{equation}
    \begin{split}
\mathcal{R}(f_{\mathcal{G}}) &= \int_{Z}\mathcal{L}_{f_{\mathcal{G}}}(z)\mathbbm{P}_{\mathcal{D}}(\mathbf{A},\mathbf{x}|y)\mathbbm{P}_{\mathcal{D}}(y)dz,\\
& = \frac{1}{2}\int_{\mathcal{X}^{+}}\mathcal{L}_{f_{\mathcal{G}}}(\mathbf{x},y)\mathbbm{P}_{\mathcal{D}}(\mathbf{A},\mathbf{x}|y=1)dx + \frac{1}{2}\int_{\mathcal{X}^{-}}\mathcal{L}_{f_{\mathcal{G}}}(x,y)\mathbbm{P}_{\mathcal{D}}(\mathbf{A},\mathbf{x}|y=-1)dx \\
&\vcentcolon= \frac{1}{2}(\mathcal{R}^{+}(f_{\mathcal{G}})+\mathcal{R}^{-}(f_{\mathcal{G}})).
    \end{split}
\end{equation}

The empirical risk tends not to approximate the expected risk due to the inconsistency of $\mathbbm{P}_{\hat{\mathcal{D}}}(\mathbf{A},\mathbf{x}|y)$ and $\mathbbm{P}_{\mathcal{D}}(\mathbf{A},\mathbf{x}|y)$. However, by introducing the error distribution with the confident pace
distribution and denoting by $\mathbb{E}_{\hat{Q}_{\lambda}}(f_{\mathcal{G}})$ in the error analysis,
we can the following error decomposition:
\begin{equation}
    \begin{split}
    &\frac{1}{2}(\mathcal{R}^{+}(f_{\mathcal{G}})+\mathcal{R}^{-}(f_{\mathcal{G}})) -\frac{1}{2}(\mathcal{R}^{+}_{emp}(f_{\mathcal{G}})+\mathcal{R}^{-}_{emp}(f_{\mathcal{G}})) \\
    & = \frac{1}{2}[\mathcal{R}^{+}(f_{\mathcal{G}}) - \mathbb{E}_{\hat{Q}_{\lambda}^{+}}(f_{\mathcal{G}}) + \mathbb{E}_{\hat{Q}_{\lambda}^{+}}(f_{\mathcal{G}})- \mathcal{R}^{+}_{emp}(f_{\mathcal{G}})] \\
    &+\frac{1}{2}[\mathcal{R}^{-}(f_{\mathcal{G}}) - \mathbb{E}_{\hat{Q}_{\lambda}^{-}}(f_{\mathcal{G}}) + \mathbb{E}_{\hat{Q}_{\lambda}^{-}}(f_{\mathcal{G}})- \mathcal{R}^{-}_{emp}(f_{\mathcal{G}})]\\
    &:= S_1 + S_2.
    \end{split}
\end{equation}

Let $S_1 = A_1 + A_2$ and $S_2 = B_1 + B_2$, where $A_1 = \frac{1}{2}(\mathcal{R}^{+}(f_{\mathcal{G}})) -\mathbb{E}_{\hat{Q}_{\lambda}^{+}}(f_{\mathcal{G}}))$, $A_2 = \frac{1}{2}(\mathbb{E}_{\hat{Q}_{\lambda}^{+}}(f_{\mathcal{G}}) -\mathcal{R}^{+}_{emp}(f_{\mathcal{G}}))$, $B_1 = \frac{1}{2}(\mathcal{R}^{-}(f_{\mathcal{G}})) -\mathbb{E}_{\hat{Q}_{\lambda}^{-}}(f_{\mathcal{G}}))$, $B_2 = \frac{1}{2}(\mathbb{E}_{\hat{Q}_{\lambda}^{-}}(f_{\mathcal{G}}) -\mathcal{R}^{-}_{emp}(f_{\mathcal{G}}))$. Here, $\mathbb{E}_{\hat{Q}_{\lambda}^{+}}(f_{\mathcal{G}})$ and $\mathbb{E}_{\hat{Q}_{\lambda}^{-}}(f_{\mathcal{G}})$ denote the pace risk with respect to positive nodes and negative nodes, respectively.

By the fact, the 0-1 loss is bounded by 1, we have:
\begin{equation} \label{A}
    \begin{split}
A_1 + A_2 = & \frac{1}{2}[\mathcal{R}^{+}(f_{\mathcal{G}}) - \mathbb{E}_{\hat{Q}_{\lambda}^{+}}(f_{\mathcal{G}}) + \mathbb{E}_{\hat{Q}_{\lambda}^{+}}(f_{\mathcal{G}})- \mathcal{R}^{+}_{emp}(f_{\mathcal{G}})] \\
&\leq \frac{1}{2}\int_{\mathcal{X}_+}(\mathbbm{P}_{\mathcal{D}}(\mathbf{A},x|y)-\hat{Q}_{\lambda}^{+}(x))dx + \frac{1}{2}{\mathfrak{R}}_{m_+}(\mathcal{H}) + \frac{1}{2}\sqrt{\frac{\ln(\frac{1}{\delta})}{2m_{+}}}\\
&\leq (1-\alpha_{\lambda})\sqrt{1-\exp{\{-D_{KL}(\mathbbm{P}^{+}_{\mathcal{D}}||E^{+})\}}} + \frac{1}{2}{\mathfrak{R}}_{m_+}(\mathcal{H}) + \frac{1}{2}\sqrt{\frac{\ln(\frac{1}{\delta})}{2m_{+}}}.
    \end{split}
\end{equation}
In a similar way, we can bound:
\begin{equation} \label{B}
    \begin{split}
    B_1 + B_2 = &\frac{1}{2}[\mathcal{R}^{-}(f_{\mathcal{G}}) - \mathbb{E}_{\hat{Q}_{\lambda}^{-}}(f_{\mathcal{G}}) + \mathbb{E}_{\hat{Q}_{\lambda}^{-}}(f_{\mathcal{G}})- \mathcal{R}^{-}_{emp}(f_{\mathcal{G}})] \\
    &\leq (1-\alpha_{\lambda})\sqrt{1-\exp{\{-D_{KL}(\mathbbm{P}^{-}_{\mathcal{D}}||E^{-})\}}} + \frac{1}{2}{\mathfrak{R}}_{m_-}(\mathcal{H}) + \frac{1}{2}\sqrt{\frac{\ln(\frac{1}{\delta})}{2m_{-}}}.
    \end{split}
\end{equation}

By taking $m^{*} = \min\{m_{-},m_{+}\}$ and combine Eq.~(\ref{A}) and Eq.~(\ref{B}), we can get:
\begin{equation} 
    \begin{split}
    \mathcal{R}(f_{\mathcal{G}}) &\leq \frac{1}{2}\mathcal{R}^{+}_{emp}(f_{\mathcal{G}})+\frac{1}{2}\mathcal{R}^{-}_{emp}(f_{\mathcal{G}})\\
    &+\frac{1}{2}\hat{\mathfrak{R}}_{m_+}(\mathcal{H}) + \frac{1}{2}\hat{\mathfrak{R}}_{m_-}(\mathcal{H}) + \sqrt{\frac{\ln(\frac{2}{\delta})}{m^*}}\\
    &+(1-\alpha_{\lambda})\sqrt{1-\exp{\{-D_{KL}(\mathbbm{P}^{+}_{\mathcal{D}}||E^{+})\}}}\\
    &+(1-\alpha_{\lambda})\sqrt{1-\exp{\{-D_{KL}(\mathbbm{P}^{-}_{\mathcal{D}}||E^{-})\}}}.
    \end{split}
\end{equation}

In addition, we further get:
\begin{equation}
{\mathfrak{R}}_{m}(\mathcal{H}) \leq \hat{\mathfrak{R}}_{m}(\mathcal{H}) + \sqrt{\frac{\ln(\frac{2}{\delta})}{2m}}.
\end{equation} 
By replacing ${\mathfrak{R}}_{m}$, we complete the proof.
\end{proof}

The above established error bounds upon 0-1 loss are hard to optimize. We change the bound of Eq.(\ref{bound_01}) under the commonly utilized hinge loss $\phi(t) = (1-t)_{+}$ for $t \in \mathbbm{R}$ and finally obtain our Theorem~\ref{bound}. The above proof is according to~\citep{gong2016curriculum}.

\section{Details of Empirical Study}
\label{appendix::dataset}

\subsection{Datasets}

In our experiments, we employ seven common datasets gathered from diverse domains. The datasets are as follows: (1) Cora, CiteSeer, and Pubmed~\citep{yang2016revisiting}, which are citation networks where nodes represent documents and edges signify citations among them; (2) WikiCS~\citep{mernyei2020wiki}, comprising nodes corresponding to Computer Science articles. Edges are based on hyperlinks, and the ten classes represent different branches of the field in the Wikipedia website; (3) Facebook~\citep{rozemberczki2021multi}, with nodes representing verified pages on Facebook and edges indicating mutual likes; (4) Physics~\citep{shchur2018pitfalls}, a co-authorship graph based on the Microsoft Academic Graph. In this dataset, nodes represent authors connected by an edge if they co-authored a paper. Node features represent paper keywords for each author’s papers, and class labels indicate the most active fields of study for each author; (5) DBLP~\citep{pan2016tri}, also a citation network, where each paper may cite or be cited by other papers. The statistical information for the utilized datasets is presented in Table~\ref{table::dataset}.

\begin{table}[ht]
\begin{center}
\caption{Important statistical information of used datasets.}
\label{table::dataset}
\begin{tabular}{lccccc}
\toprule
 Dataset  &  Edges &  Classes &  Features &  Nodes/Labeled Nodes &  Labeled Ratio \\
\hline
Cora  & $5,429$ & $7$ & $1,433$ & $2,708$/$1208$ & $44.61\%$\\
CiteSeer  &$4,732$& $6$ &$3,703$ & $3,327$/$1827$ & $54.91\%$\\
PubMed  & $44,338$ & $3$ & $500$ & $19,717$/$18217$ & $92.39\%$\\
WikiCS   & $215,603$ & $10$ & $300$ & $11,701$/$580$ & $4.96\%$\\
Facebook & $342,004$ & $4$ & $128$ & $22,470$/$400$ & $1.78\%$   \\
Physics & $495,924$ & $5$ & $8415$ & $34,493$/$500$ & $1.45\%$\\
DBLP & $105,734$ & $4$ & $1639$ & $17,716$/$800$ &$4.51\%$ \\
\bottomrule
\end{tabular}
\end{center}
\vspace{-0.9em}
\end{table}

\subsection{Label Noise Generation Setting} \label{lng}

Following previous works~\citep{dai2021nrgnn,du2021pi,xia2020part}, we consider three settings of simulated noisy labels:
\begin{itemize}
\item[(1)] \textit{Symmetric} noise: this kind of label noise is generated by flipping labels in each class uniformly to incorrect labels of other classes. 
\item[(2)] \textit{Pairflip} noise: the noise flips each class to its adjacent class. More explanation about this noise setting can be found in~\citep{yu2019does,zheng2020error,lyu2019curriculum}.\item[(3)] \textit{Instance-dependent} noise: the noise is quite realistic, where the probability that an instance is mislabeled depends on its features. We follow~\citep{xia2020part} to generate this type of label noise to validate the effectiveness of the proposed method.
\end{itemize}

\subsection{Baseline Details} \label{baseline}
In more detail, we employ baselines:
\begin{itemize}
    \item \textit{Sample selection with label noise on i.i.d. data:}
    \begin{itemize}
    \item[(1)] Co-teaching+~\citep{yu2019does}: This approach employs a dual-network mechanism to reciprocally extract confident samples. Specifically, instances with minimal loss and discordant predictions are identified as reliable, clean samples for subsequent training.
    \item [(2)]  Me-Momentum~\citep{bai2021me}: The objective of this method is to identify challenging clean examples from noisy training data. This process involves iteratively updating the extracted examples while refining the classifier.
    \item [(3)] MentorNet~\citep{jiang2018mentornet}: This approach involves pre-training an additional network, which is then used to select clean instances and guide the training of the main network. In cases where clean validation data is unavailable, the self-paced variant of MentorNet resorts to a predefined curriculum, such as focusing on instances with small losses.
    \end{itemize}

    \item \textit{Graph Curriculum learning:}
    \begin{itemize}
    \item [(1)]CLNode~\citep{wei2023clnode}: CLNode is a curriculum learning framework aimed at enhancing the performance of backbone GNNs by gradually introducing more challenging nodes during the training process. The proposed difficulty measure is based on label information.

    \item [(2)] RCL~\citep{zhang2023relational}:  RCL utilizes diverse underlying data dependencies to train improved Graph Neural Networks (GNNs), resulting in enhanced quality of learned node representations. It gauges the inter-node relationships as a measure of difficulty for each node.

    \end{itemize}

\item \textit{Denoising methods on graph data:}
    \begin{itemize}
    \item [(1)] LPM~\citep{xia2020towards}: The method is specifically tailored to address noisy labels in node classification, employing a small set of clean nodes for guidance.
    \item [(2)] CP~\citep{zhang2020adversarial}: The method operates on class labels derived from clustering node embeddings. It encourages the classifier to comprehend class-cluster information, effectively mitigating overfitting to noisy labels. Prior to clustering, node embeddings are acquired using the Node2Vec model~\citep{grover2016node2vec}.
    \item [(3)] NRGNN~\citep{dai2021nrgnn}: In this approach, a label noise-resistant GNN establishes connections between unlabeled nodes and noisily labeled nodes with high feature similarity. This connection strategy effectively incorporates additional clean label information into the model.
    \item [(4)] PI-GNN~\citep{du2023noise}: This method introduces Pairwise Intersection (PI) labels, generated based on feature similarity among nodes. These PI labels are then employed to alleviate the adverse impact of label noise, thereby enhancing the model's robustness.
    \item [(6)] RS-GNN~\citep{dai2022towards}
       This method primarily aims to improve the robustness of Graph Neural Networks (GNNs) in the presence of noisy edges. It achieves this by training a link predictor on graphs with inaccuracies in edge connections, ultimately enabling GNNs to effectively learn from such imperfect graph structures.
     \item [(5)] RT-GNN~\citep{qian2023robust}: This approach identifies clean labeled nodes by leveraging the memorization effect of neural networks. Subsequently, it generates pseudo-labels based on these selected clean nodes to mitigate the impact of noisy nodes during the training process.
    \end{itemize}

\end{itemize}

\subsection{Algorithm Framework of TSS}

\begin{algorithm}[H]
\caption{Algorithm flow of TSS.}
\label{alg:TSS}
\begin{algorithmic} [1]
\STATE { \bfseries Input:} A pretrained classifier $f_{\mathcal{G}}^{p}$, the noisy training set $\tilde{\mathcal{D}}_{\mathrm{tr}}=\{(\mathbf{A},x_i,\tilde{y}_i)\}^{n_{\mathrm{tr}}}_{i=1}$, the identity matrix $\mathbf{I}$, the normalized adjacency matrix $\hat{\mathbf{A}}$, the hyperparameters $\alpha$, $\lambda_{0}$, $T$ 
\STATE  Obtain $\boldsymbol{\pi} \leftarrow \alpha(\mathbf{I}-(1-\alpha)\hat{\mathbf{A}})^{-1} $
\STATE Initialize parameters of a GNN classifier $f_{\mathcal{G}}$
\STATE Let $t = 1$
\WHILE{$t < T$ or not converge}
    \FOR{$\mathbf{v}_i \in \tilde{\mathcal{D}}_{\mathrm{tr}}$}
    \STATE Calculate $\mathbf{Cb}_{i} \leftarrow \frac{1}{n_{\mathrm{tr}}(n_{\mathrm{tr}}-1)}\sum_{\substack{\mathbf{v}_u \neq\mathbf{v}_i \neq\mathbf{v}_v \\ \tilde{y}_u \neq \tilde{y}_v}}
    \frac{\boldsymbol{\pi}_{u,i}\boldsymbol{\pi}_{i,v}}{\boldsymbol{\pi}_{u,v}}$
    \ENDFOR
\STATE Sort $\tilde{\mathcal{D}}_{\mathrm{tr}}$ according to $\mathbf{Cb}_{i}$ in ascending order
\STATE $\lambda_{t} \leftarrow \min(1, \lambda_{t-1} + (1-\lambda_{t-1})*\frac{t}{T})$
\STATE Generate noisy training subset $\tilde{\mathcal{D}}_{\mathrm{tr}}^{t} \leftarrow \tilde{\mathcal{D}}_{\mathrm{tr}}[1,\dots,\lfloor \lambda_{t}*n_{\mathrm{tr}} \rfloor]$
\STATE Extract confident training subset $\hat{\mathcal{D}}_{\mathrm{tr}}^{t}$ from $\tilde{\mathcal{D}}_{\mathrm{tr}}^{t}$
\\ \textcolor{gray}{// i.e., the training nodes whose noisy labels are identical to the ones predicted by $f_\mathcal{G}^{p}$}
\STATE Calculate loss $\mathcal{L}$ on $\hat{\mathcal{D}}_{\mathrm{tr}}^{t}$
\STATE Back-propagation on $f_{\mathcal{G}}$ for minimizing $\mathcal{L}$
\STATE $t \leftarrow t +1 $
\ENDWHILE
\STATE {\bfseries Output:}Trained GNN classifier $f_{\mathcal{G}}$ 
\end{algorithmic}
\end{algorithm}

\subsection{Pacing Function of TSS}

After measuring node difficulty using the CBC measure, we employ the TSS method to enhance the training of our GNN model. We incorporate a pacing function $\lambda(t)$ to govern the proportion $\lambda$ of training nodes available at the $t$-th epoch. In TSS, we utilize three distinct pacing functions: linear, root, and geometric. 

\begin{itemize}
    \item linear:
    \begin{equation}
        \lambda_{t} = \min(1, \lambda_{t-1} + (1-\lambda_{t-1})*\frac{t}{T})
    \end{equation}
    \item root:
    \begin{equation}
        \lambda_{t} = \min(1, \sqrt{\lambda_{t-1}^{2} + (1-\lambda_{t-1}^{2})*\frac{t}{T}})
    \end{equation}
    \item geometric:
    \begin{equation}
        \lambda_{t} = \min(1, 2^{\log_{2}\lambda_{t}-\log_2\lambda_{t}*\frac{t}{T}})
    \end{equation}
\end{itemize}

The linear function escalates the training node difficulty uniformly over epochs. On the other hand, the root function introduces a higher proportion of difficult nodes in a smaller number of epochs. Meanwhile, the geometric function extends the training duration on a subset of easy nodes by conducting multiple epochs.

\subsection{Implementation Details} 
A two-layer graph convolutional network whose hidden dimension is 16 is deployed as the backbone for all methods. We apply an Adam optimizer~\citep{kingma2014adam} with a learning rate of $0.01$. The weight decay is set to $5\times10^{-4}$. The number of pre-training epochs is set to 400. While the number of retraining epochs is set to 500 for Cora, CiteSeer, and 1000 for Pubmed, WikiCS, Facebook, Physics and DBLP. All hyper-parameters are tuned based on a noisy validation set built by leaving 10\% noisy training data.

\section{More experiment}
\label{app:E}

\subsection{Visualize extracted nodes in TSS}

\begin{figure*}
  \centering 

{
    \begin{minipage}[b]{0.25\linewidth}
      \centering
      \includegraphics[width=\linewidth]{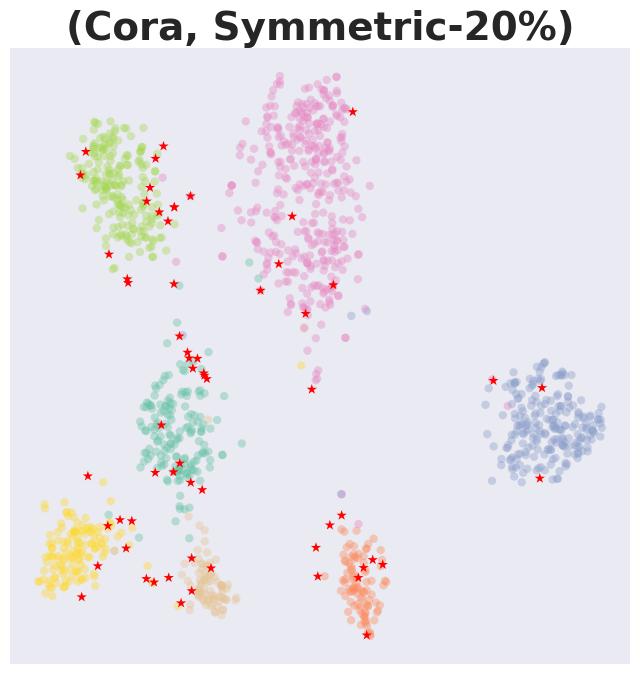}
      \includegraphics[width=\linewidth]{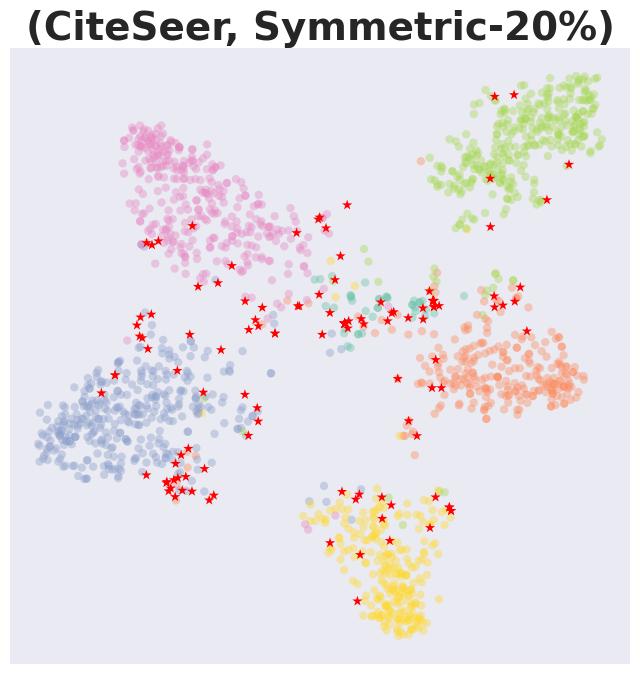}
    \end{minipage}
  }
\hspace{-2mm}
{
    \begin{minipage}[b]{0.25\linewidth}
      \centering
      \includegraphics[width=\linewidth]{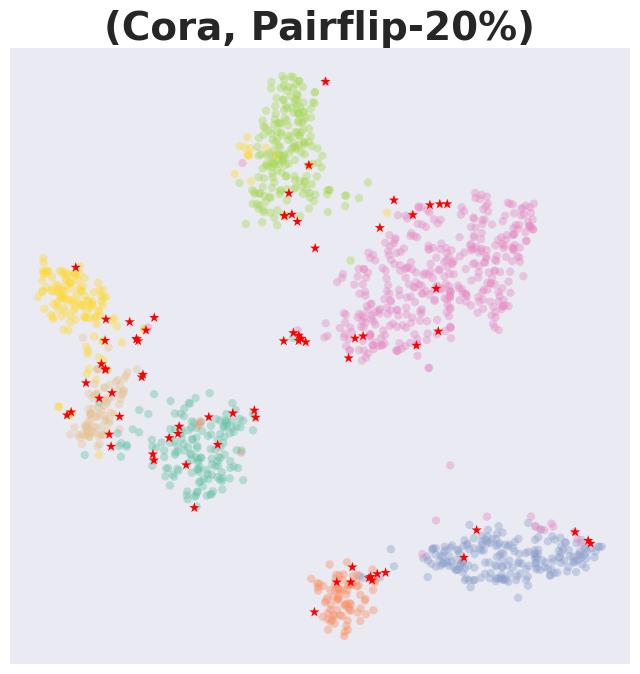}
      \includegraphics[width=\linewidth]{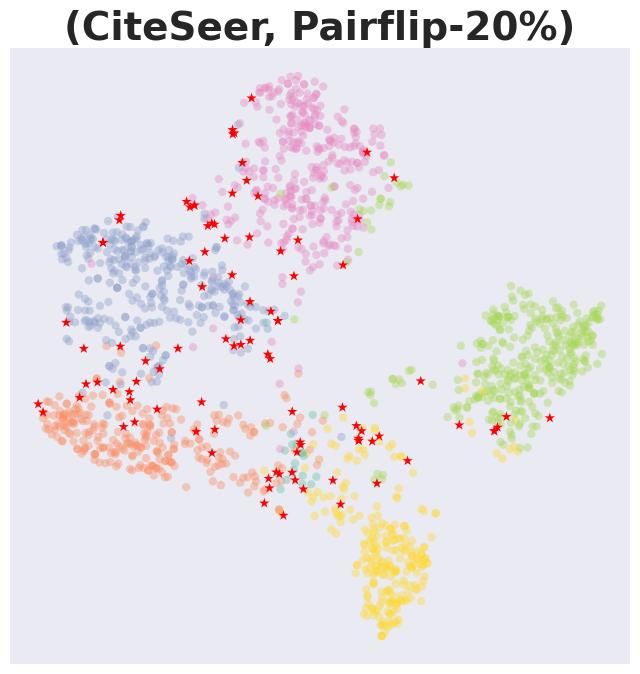}
    \end{minipage}
    }
\hspace{-2mm}
{
    \begin{minipage}[b]{0.25\linewidth}
      \centering
      \includegraphics[width=\linewidth]{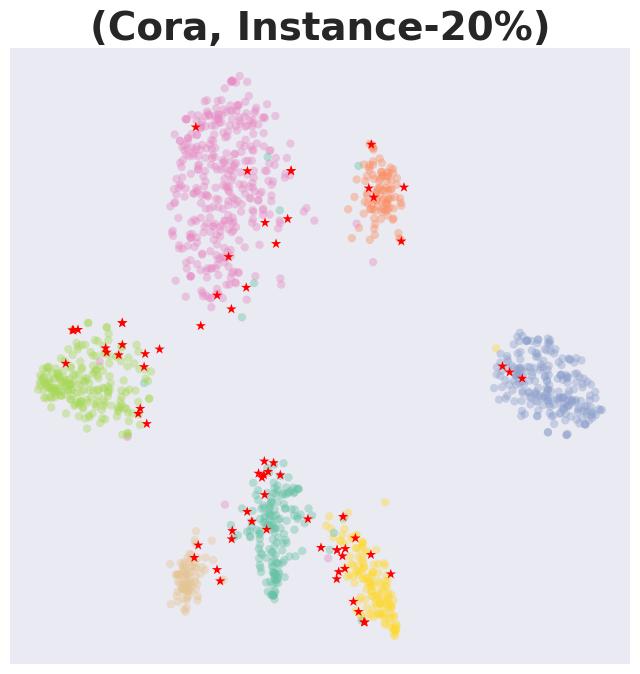}
      \includegraphics[width=\linewidth]{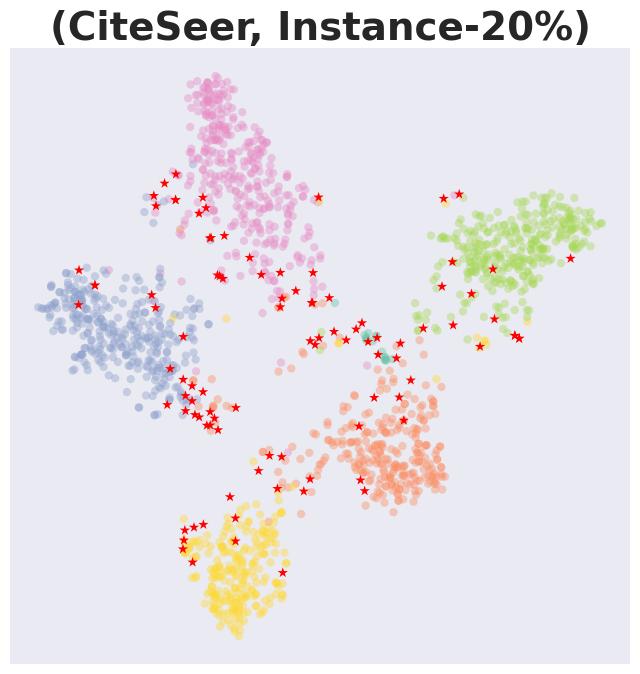}
    \end{minipage}
  }

  \vfill
  \caption{Visualization of the extracted nodes in the TSS. The red dots represent the newly extracted nodes in the later stage. Other colour dots represent the nodes extracted in the early stage.}
  \label{fig:Visualization}
\end{figure*}

To justify that TSS can extract clean near-boundary nodes, we visualize the extracted clean nodes by employing t-SNE~\cite{van2008visualizing} on their embeddings, which are the penultimate layer representation vectors. The results are shown in Figure~\ref{fig:Visualization}, where red dots represent the nodes extracted in the later stage of TSS and other colour dots represent the nodes extracted in the early stage of TSS. On Cora and CiteSeer, we can clearly see that there are lots of red dots which are on the boundary of class-clusters. This supports and justifies our claim that TSS can extract clean informative nodes (located on a class's boundary and link the nodes of different classes). Comparing the nodes extracted on three types of noisy datasets, we can observe that the TSS is not sensitive to the type of label noise and can work well on the most general instance-dependent label noise cases.

\subsection{CBC distributions of nodes with varying homophily ratio}

\begin{figure*} [h]
\centering
\includegraphics[width=1\linewidth]{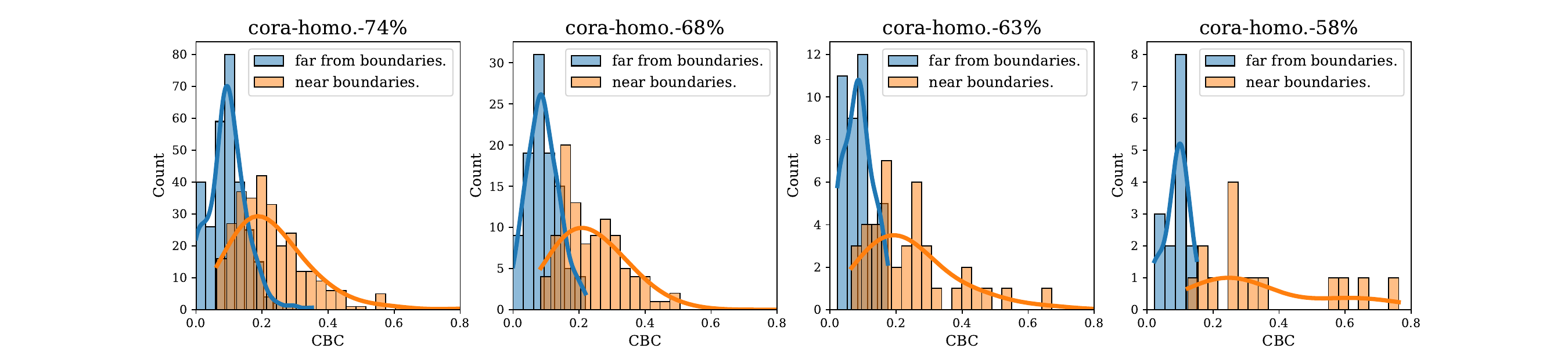}
\caption{The distributions of the CBC score \textit{w.r.t.} nodes on CORA with different homophily ratios in the presence of 30\% instance-dependent label noise. The nodes are considered ``far from topological class boundaries" (far from boundaries.) when their two-hop neighbours belong to the same class; conversely, nodes are categorized as ``near topological class boundaries" (near boundaries.) when this condition does not hold.}
\label{figure::CBC_distribution_homo}
\end{figure*}

In this section, we assess the effectiveness of our CBC measure in relation to varying homophily ratios within the noisy labeled graph. We modify the graph structure by introducing synthetic, cross-label (heterophilous) edges that connect nodes with differing labels. The methodology for adding these heterophilous edges, as well as the calculation for the homophily ratio, are both referred to~\citep{ma2021homophily}. As illustrated in Fig.~\ref{figure::CBC_distribution_homo}, a decrease in the homophily ratio results in an increased number of nodes near class boundaries, which consequently exhibit higher CBC scores. Notably, our CBC measure effectively reflects the topology of nodes even as the complexity of the graph increases.

\vspace{-5pt}

\subsection{Performance comparison on heterphily datasets}

We evaluate the effectiveness of our method on three commonly used heterogeneous datasets, i.e., DBLP~\citep{fu2020magnn}, Chameleon~\citep{rozemberczki2021multi}, Squirrel~\citep{rozemberczki2021multi} under $30\%$ instance-dependent label noise. The summary of experimental results is in the Table~\ref{tab:heter_data}. As can be seen, our method still shows superior performance over a range of baselines. 

\begin{table*}
    \caption{Mean and standard deviations of classification accuracy (percentage) on heterphily graph datasets with 30$\%$ instance-dependent label noise. The results are the mean over five trials and the best are bolded.}
	\centering
	  \footnotesize
	\setlength\tabcolsep{3pt}
	{
         \midsepremove
	\begin{tabular}{l |c|c|c}
		\toprule
             Method & {\textit{Chameleon}} & {\textit{Squirrel}} & {\textit{DBLP}}\\
            \hline

		\hline	

		CP  & 55.08±2.18 & 43.42±2.46 &  70.02±3.06 \\
  
  		NRGNN & 	49.02±2.35 & 41.35±1.98 &	  72.48±2.61 	\\

             PI-GNN & 52.85±2.16 & 43.31±2.97	  &  	71.72±3.39 \\

		Co-teaching+ & 53.07±1.98 & 39.48±2.54  &   66.32±2.12 \\

		Me-Momentum  & 55.01±1.69 & 44.38±1.78 &   59.88±0.60 \\
    
     MentorNet & 53.73±3.75 & 39.63±3.43  &   63.73±4.93  \\

      CLNode & 52.85±2.91 & 35.92±1.84  &   72.32±2.06  \\

        RCL &52.96±0.96 & 40.59±1.23  &   63.20±0.81  \\

\hline
         \rowcolor{LightGreen}
        TSS & \textbf{56.17±0.28}	& \textbf{48.03±1.03} &  \textbf{74.70±1.72} \\

		\bottomrule
	\end{tabular} 
 }
	\label{tab:heter_data}
\end{table*}

\section{Limitations}
Indeed, our TSS method has demonstrated effectiveness across various scenarios. However, it's important to acknowledge certain inherent limitations due to the intricacies of dealing with noisily labeled graphs.

Firstly, the TSS method is specifically tailored for homogeneously-connected graphs, where linked nodes are anticipated to share similarities. This is evident in the diverse datasets utilized in our experiments. Adapting TSS to heterogeneously connected graphs, such as protein networks, requires a nuanced refinement of our approach to suit the distinct network characteristics.

Secondly, a notable challenge for TSS arises when the labeling ratio is exceptionally low. In such instances, the extraction of clean nodes might inadvertently overlook crucial features of mislabeled nodes. This oversight could potentially impact the learning process of models. Addressing this limitation mandates thoughtful adjustments in our approach, aiming to accommodate scenarios with scantily labeled data better.

\end{document}